\documentclass[conference]{IEEEtran}

\usepackage{tikz}
\usepackage{amsmath}

\usepackage{filecontents}

\usepackage[font=bf]{caption}
\usepackage{times}
\usepackage{subfig}
\usepackage{xspace}
\usepackage{graphicx}
\usepackage{clrscode}
\usepackage{breakurl}

\usepackage{hyperref}

\usepackage{mathtools}
\usepackage{amsmath}
\usepackage{amsfonts}

\usepackage{multicol}

\newcounter{packednmbr}

\newenvironment{packeditemize}{
\begin{list}{$\bullet$}{
\setlength{\itemsep}{1.5pt}
\setlength{\labelwidth}{8pt}
\setlength{\leftmargin}{10pt}
\setlength{\labelsep}{3pt}
\setlength{\listparindent}{\parindent}
\setlength{\parsep}{1.5pt}
\setlength{\parskip}{1.5pt}
\setlength{\topsep}{1.5pt}}}{\end{list}}

\newcommand{\tightcaption}[1]{\vspace{-0pt}\caption{#1}\vspace{-0pt}}
\newcommand{\tightsection}[1]{\vspace{-0in}\section{#1}\vspace{-0cm}}
\newcommand{\tightsubsection}[1]{\vspace{-0in}\subsection{#1}\vspace{-0cm}}

\newcommand{\ty}[1]{{\footnotesize\color{red}[TY: #1]}}
\newcommand{\Commentout}[1]{}

\usepackage{footnote}
\usepackage{pifont}
\newcommand{\Name}{\textsc{REFN}\xspace}

\newcommand{\nday}{1-day/n-day\xspace}
\newcommand{\netpatch}{VNF fixes\xspace}

\newcommand{\manualpatch}{manual patching\xspace}
\newcommand{\softwarepatch}{patch management software\xspace}
\newcommand{\gmlpatch}{generic ML-based patching\xspace}
\newcommand{\gllmpatch}{generic LLM-based patching\xspace}
\newcommand{\networkfiltering}{manual network filtering\xspace}
\newcommand{\gmlfiltering}{generic ML-based network filtering\xspace}
\newcommand{\gllmfiltering}{generic LLM-based network filtering\xspace}

\newcommand{\Distillation}{Agentic-RAG-based Knowledge Distillation\xspace}
\newcommand{\RLVNF}{RL-From-VNF Pipeline\xspace}
\newcommand{\Validator}{Online Agentic Validator\xspace}

\newcounter{painnmbr}
\newcounter{painlabel}
\renewcommand{\thepainlabel}{\textbf{\thepainnmbr}}

\usepackage{listings}
\usepackage{amsfonts}
\usepackage{times}
\usepackage{xspace} 
\usepackage{epsfig} 
\usepackage{amsmath}
\usepackage{amssymb}
\usepackage{fancyvrb}
\usepackage{graphicx}
\usepackage{subfig}
\usepackage[english]{babel}
\usepackage{courier}
\usepackage{clrscode}
\usepackage{pbox}
\usepackage{multirow}

\usepackage{makecell}

\usepackage{algorithm}
\usepackage{algorithmic}

\begin{document}

\title{\Name: A Reinforcement-Learning-From-Network \\ 
Framework against \nday Exploitations}



%

\author{
\IEEEauthorblockN{
		Tianlong Yu\textsuperscript{1},
            Lihong Liu\textsuperscript{1},
		Ziyi Zhou\textsuperscript{1},
            Fudu Xing\textsuperscript{3},
		Kailong Wang\textsuperscript{2},
		Yang Yang\textsuperscript{1}} \\ 
	\IEEEauthorblockA{\textsuperscript{1}School of Artificial Intelligence, Hubei University, Wuhan, China\\}
    \IEEEauthorblockA{\textsuperscript{2}Huazhong University of Science and Technology, Wuhan, China\\}
    \IEEEauthorblockA{\textsuperscript{3}University of Southern California, Los Angeles, USA\\}
    tommyyu21@163.com, llh3401939433@outlook.com, 202421121013087@stu.hubu.edu.cn, \\ fuduxing@usc.edu, wangkl@hust.edu.cn, yangyang@hubu.edu.cn
}




\maketitle

\begin{abstract}

The exploitation of \nday vulnerabilities poses severe threats to networked devices due to massive deployment scales and delayed patching (average Mean-Time-To-Patch exceeds 60 days). Existing defenses, including host-based patching and network-based filtering, are inadequate due to limited \textit{scalability} across diverse devices, \textit{compatibility} issues especially with embedded/legacy systems, and \textit{error-prone} deployment process (e.g., manual patch validation).
To address these issues, we introduce REFN (Reinforcement-Learning-From-Network), a novel framework that trains Large Language Models (LLMs) to autonomously generate network filters to prevent \nday exploitations.
%
%
%
\Name ensures scalability by uniquely employs Reinforcement Learning (RL) driven by online network rewards instead of traditional Human Feedback (RLHF).
\Name guarantees compatibility via unified deployment on edge security gateways (e.g., Amazon Eero).
\Name provides robustness via online validation using real network traffic. 
Crucially, \Name addresses three core challenges in \textit{training LLMs for exploit prevention}:
1) expanding current LLMs' \textit{limited vulnerability-fixing expertise} via Agentic-RAG-based Knowledge Distillation;
2) bridging current LLMs' \textit{language-to-network gaps} through an RL-From-VNF Pipeline that translates language context (e.g., vulnerability description) into network enforcement;
3) addressing the \textit{LLM hallucination \& non-determinism} via the Online Agentic Validation that penalizes erroneous outputs.
Evaluated across 22 families of \nday exploits,
\Name demonstrates \textit{effectiveness} (21.1\% higher accuracy than alternatives), \textit{efficiency} (Mean-Time-To-Patch of 3.65 hours) and \textit{scalability} (easily scale to 10K devices).
\Name serves as an initial step toward training LLMs to rapidly prevent massive-scale \nday exploitations.

\end{abstract}

\tightsection{Introduction}
\label{sec:introduction}

The real-world conflicts are rapidly expanding into cyberspace, driving massive-scale exploitations of \nday vulnerabilities across networked devices~\cite{Alanazi2023LoadOA, log4jnews}. 
Landmark incidents, such as the 2021 Log4j vulnerability which affecting hundreds of millions of devices~\cite{log4jnews}, demonstrate the severity of this threat. 
Offensive capabilities are further amplified by LLM-powered tools (e.g., WormGPT~\cite{wormgpt}, HackerGPT~\cite{hackergpt}), while defenses are slow to respond and falling behind, e.g., critical patches face dangerous delays, with Mean-Time-To-Patch (MTTP) averaging 60 to 150 days~\cite{mttp}.

Current vulnerability fixing strategies primarily utilize two approaches: host-based patching~\cite{patchmanager, chocolatey, avira, ninite, patchupdater, sumo, heimdal, npackd, ruckzuck, patchrnn, graphspd, pavudi, spi, inferfix} and network-based filtering~\cite{kitsune, snort, odds, wang2004shield, chatgpt, mscopilot, llama}. Host-based patching operates by updating software or hardware on vulnerable devices through patch generation, installation, and validation. 
Despite automation efforts including patch management tools~\cite{patchmanager, chocolatey, avira, ninite, patchupdater, sumo, heimdal, npackd, ruckzuck} (largely restricted to standard computers) and more recent generic ML-based approaches~\cite{patchrnn, graphspd, pavudi, spi, inferfix}, host-based patching solutions remain challenged by source code availability and the prohibitive cost of upgrading embedded/legacy systems.
Network-based filtering includes manual rule-based filtering~\cite{snort} (inherently error-prone and not scalable), generic ML-based network filtering~\cite{kitsune, odds} (relying on statistical anomaly detection which causes disruptive false positives and fails against low-frequency attacks like APTs).
It is also worth noting that generic LLMs can be used to generate patches or filtering rules via manual prompts.
However, the generic LLMs~\cite{chatgpt, mscopilot, llama} are plagued by hallucination and non-determinism, generating seemingly correct but functionally flawed patches and filtering rules.

Current vulnerability fixing mechanisms are ill-equipped to address large-scale 1-day/n-day exploitation targeting diverse networked infrastructure. Consider the compromise of millions of smart meters within an Advanced Metering Infrastructure (AMI) vulnerable to Smart Grid attacks like load oscillation~\cite{Alanazi2023LoadOA}. 
Fixing such a vulnerability via host-based patching demands expert analysis and expensive deployment/verification across millions of devices. 
Network-based filtering alternatives, which rely on manual rules or anomaly detection, similarly fail to scale effectively to millions of AMI endpoints while introducing unacceptable risks of disrupting critical grid operations through false positives.
We argue that existing approaches fail to address large-scale 1-day/n-day exploits effectively due to three fundamental limitations:

\begin{packeditemize}
\item \textbf{Scalability}: 
Patch and filtering rule generation relies critically on domain experts to manually analyze code and craft fixes—a process that is inherently unscalable. This is acutely demonstrated by vulnerabilities like Log4j, which impacted millions of heterogeneous devices from Apache servers to Smart City cameras. Manually generating tailored fixes for such diversity is prohibitively slow and costly.

    
\item \textbf{Compatibility}:
Host-based patching suffers severe compatibility issues, particularly for embedded or legacy systems. A Log4j patch designed for Windows servers, for instance, will typically fail or malfunction on resource-constrained smart cameras due to platform and dependency mismatches.
    
    
\item \textbf{Error-susceptibility}: 
Both patching and filtering approaches are intrinsically susceptible to errors, as validation remains exceptionally challenging across diverse device functionalities. 
For example, it is hard to verify that an AMI smart meter’s firmware patch installs correctly and will not disrupting critical grid operations.
    
\end{packeditemize}

To overcome these limitations, we introduce \Name (Reinforcement-Learning-From-Network):
a novel framework that trains Large Language Models (LLMs) to generate network filters that can be deployed on edge security gateways to prevent 1-day/n-day exploits. 
REFN's core innovation include leveraging Reinforcement Learning (RL) driven by real-time network rewards – not human feedback (RLHF), enabling autonomous adaptation to evolving \nday threats.
\Name ensures scalability as its RL-trained LLMs automatically generate tailored filters, eliminating dependency on manual patch/fix generation.
\Name guarantees compatibility via unified deployment on edge security gateways (e.g., Amazon Eero) abstracts away device-specific complexities, ensuring broad coverage including embedded/legacy systems.
\Name is error-unsusceptible, as it can continuously perform online validation against real network traffic, and can correct erroneous filters by adjusting the online reward generated from network dataplane.



In developing \Name, we address three core challenges in \textit{training LLMs for exploit prevention}: 
1) \textbf{Limited vulnerability-fixing expertise}: LLMs trained on general-purpose data lack internalized vulnerability-fixing inference capabilities (e.g., unaware of day-1 vulnerability);
2) \textbf{Language-to-network gaps}: LLMs optimized for natural language (e.g., via RLHF) struggle to translate textual descriptions into executable network enforcements due to structural mismatches between linguistics and network;
3) \textbf{LLM hallucination and non-determinism}: LLMs generate inconsistent or erroneous outputs across iterations, producing unreliable filtering rules that disrupts normal functions.





    


To address these challenges, REFN introduces three novel designs: 
1) \textbf{Agentic-RAG-based Knowledge Distillation}: dynamically retrieves and internalizes vulnerability intelligence from security databases, CVE reports, and historical fixes, enabling contextual reasoning for precise filter generation.   
2) \textbf{RL-from-VNF Pipeline}: employs reinforcement learning (RL) driven by real-time rewards from Virtualized Network Function (VNF) to translate textual vulnerability descriptions into protocol-aware network enforcements.
3) \textbf{Online Agentic Validation}: enforces output reliability through real-time network feedback loops that penalize errors and refine filters during deployment.

The main contributions of this paper are as follows:

\begin{packeditemize}

    \item \textbf{Proof-of-Concept}: 
    We demonstrate the viability of training LLMs for preventing massive-scale \nday exploits. 

    \item \textbf{\Name Framework}: 
    We introduce a novel framework that leverages Reinforcement
Learning (RL) driven by real-time network rewards – instead of
human feedback (RLHF), to generate vulnerability-fixing filters. 
REFN is scalable, compatible and error-unsusceptible. Code available:  \url{https://github.com/REFN2025/REFN2025}.

    \item \textbf{Security-Specialized LLM Model}: We provide the RL-trained vulnerability-fixing LLM specifically effective against 22 families of \nday exploits. Model available:  \url{https://huggingface.co/llhview/HuggingFace/tree/main}.
    
    \item \textbf{RL Dataset for Exploit Prevention}: We present the first dataset enabling Reinforcement Learning of LLMs to prevent \nday exploits, covering 22 families of exploits and 65 types of devices. Dataset available:  
    \url{https://github.com/REFN2025dataset/REFN2025/tree/master}.

    \item \textbf{Rigorous Evaluation}: Using our RL dataset, we comprehensively evaluate \Name and demonstrate its \textit{effectiveness} (21.1\% higher accuracy than alternatives), its \textit{efficiency} (Mean-Time-To-Patch of 3.65 hours) and its \textit{scalability} (easily scale to 10K devices).

\end{packeditemize}

\tightsection{Motivations and Related Works}
\label{sec:relatedworks}


\subsection{Current vulnerability fixing approaches}

Current vulnerability fixing mechanisms, including host-based patching and host-based patching,
can be further specified as several types of approaches:


%
%

\noindent\textbf{Manual patching}:
This approach involves human-driven updates to software/hardware systems through three core phases: patch generation, installation, and validation. 
Security administrators must manually execute each step, requiring substantial effort and expertise while offering limited scalability. 


\noindent\textbf{Patch management software}:
Developed to automate patching workflows~\cite{patchmanager, chocolatey, avira, ninite, patchupdater, sumo, heimdal, npackd, ruckzuck}, these tools primarily streamline delivery and installation for standard computing environments. However, they are ineffective for embedded or legacy systems due to platform-specific dependencies and lack of validation.


\noindent\textbf{Generic ML-based patching}:
InferFix~\cite{inferfix} employs ML-assisted repair; GraphSPD~\cite{graphspd} and PAVUDI~\cite{pavudi} utilize GNNs for patch analysis; PatchRNN~\cite{patchrnn} models patches via RNNs; SPI~\cite{spi} identifies patches through commits. 
ML solutions~\cite{patchrnn, graphspd, pavudi, spi, inferfix} are constrained by source code availability and face prohibitive deployment costs for legacy/embedded devices.


\noindent\textbf{Manual network filtering}:
This approach requires the security admins to craft rule-based filters or access control rules manually~\cite{he2018contextpolicy, contexiot},
inherently limited by human error and impractical for large-scale deployments. The lack of automation also impedes rapid response to emerging threats.


\noindent\textbf{Generic ML-based network filtering}:
These methods detect threats through statistical traffic anomalies~\cite{kitsune, odds, bartos2016optimized, flowlens, fu2021realtime, deeplog, jaqen}: Kitsune~\cite{kitsune} uses autoencoder reconstruction deviations; Bartos et al.~\cite{bartos2016optimized} and FlowLen~\cite{flowlens} apply flow-sampling classifiers; Whisper~\cite{fu2021realtime} analyzes periodicity via spectral decomposition; DeepLog~\cite{deeplog} models system logs. Despite sophistication, they remain vulnerable to low-frequency APT evasion due to reliance on pattern deviations.


\noindent\textbf{LLM based patching/filtering}:
By prompting generic LLMs~\cite{chatgpt, mscopilot, llama}, this approach generates patches or filtering rules. However, it suffers from critical hallucination and non-determinism issues, producing seemingly valid but functionally flawed outputs that compromise security efficacy.



\subsection{Issues with current approaches}

\textbf{Scalability issue}:
Current vulnerability remediation mechanisms face significant scalability challenges. The sheer diversity of vulnerable devices and systems—encompassing complex infrastructures, numerous software applications, and millions of endpoints—makes broad protection difficult. A prime example is the Log4j vulnerability, which impacted devices ranging from Apache servers to consumer appliances like Siemens refrigerators. Manually generating and validating patches or filtering rules for each unique vulnerability-device combination demands immense domain expertise and is impractical at scale. Furthermore, patch deployment itself consumes substantial time, manpower, and technical resources, a burden particularly acute for organizations with limited budgets. This scalability gap is exacerbated by the emergence of LLM-empowered exploitation tools (e.g., HackerGPT~\cite{hackergpt}, WormGPT~\cite{wormgpt}), which dramatically enhance attackers' ability to launch large-scale exploits.


\textbf{Compatibility issue}:
Existing approaches also struggle with compatibility across diverse device ecosystems. End-of-life devices often lack manufacturer support and receive no security updates, leaving them persistently vulnerable. Many older or resource-constrained devices lack the hardware or software capability to accept firmware updates at all. Proprietary, closed-source systems present another barrier, as they cannot be audited by the community or have third-party patches developed. Furthermore, patching complex hardware like medical equipment or industrial control systems frequently requires specialized expertise that may be unavailable.


\textbf{Error-susceptibility issue}:
Current patching mechanisms are inherently susceptible to errors. Fixes often require modifications across multiple system components, increasing the risk of implementation mistakes that can cause instability or functional loss. Conflicts with existing software or hardware versions are common, and managing dependencies adds further complexity. Crucially, the process relies heavily on manual intervention for both deployment and validation, introducing significant opportunities for human error. It is inherently challenging to anticipate all potential issues—especially in complex systems or embedded devices—meaning even validated patches can inadvertently introduce new problems or unexpected behavior. For example, the ML-based network filtering approaches
may flag the benign accesses on a newly joined devices as malicious as it is rarely seen deviation.


\subsection{New vantage point to prevent \nday vulnerabilities}

To tackle the \textit{compatibility} and \textit{scalability} issue, security vendors are shifting the vulnerability fixing function from host-side to 
\textit{Edge Security Gateways (ESG)},
including Amazon eero~\cite{amazoneero}, Cisco Meraki~\cite{ciscomeraki}, Netgear Orbi~\cite{netgearorbi} and Linksys Velop~\cite{linksysvelop}.
In such network-fix paradigm, the vulnerability fixing is enforced as network filtering on the edge security gateways. 
The remote cloud services is responsible for generating the filtering rules and installing them on the edge security gateways.
For example, the Cisco Talos Intelligence cloud service can generate a network filtering for Log4j, and deploy it on Meraki MX edge routers to detect and block Log4j exploits~\cite{merakilog4jint,merakilog4jpatch}.
The gateways hosting the vulnerability fixes are unified platforms
such as Cisco IOS~\cite{ciscoios}, Rasberry PI~\cite{rasberrypi} or OpenWRT~\cite{openwrt}.
The network-based patches only need to adapt to several unified edge platforms instead of heterogeneous vulnerable devices.
Unlike current host-based patching mechanisms - which are hard to upgrade and slow to respond to emerging vulnerabilities, the network-fix update~\cite{khalid2019correctness} can be performed in seconds.

\section{Overview}
\label{sec:overview}

In this section, we present the overview of \Name - a scalable, compatible and error-unsusceptible framework that trains Large Language Models (LLMs) to autonomously generate and deploy network filters on Edge Security Gateways (ESG), and prevent \nday exploitations across heterogeneous networked devices. 



\textbf{Threat model}:
Before detailing REFN's design, we establish the following assumptions regarding adversarial capabilities and system trust boundaries:

\textit{Adversarial Capabilities}: 1) The adversary can exploit all \nday vulnerabilities in exposed devices; 2) The adversary can utilize LLM-based automation tools (e.g., HackerGPT~\cite{hackergpt}, WormGPT~\cite{wormgpt}) to generate and scale exploits.

\textit{System Trust Assumptions}: 1) The Edge Security Gateway (ESG) is a secure, trusted component and cannot be compromised by the attacker;
    2) Cloud servers responsible for training the LLMs, generating and deploying network filters onto the edge security gateways are trusted and secure;
    3) The edge security gateways have the capability to enforce network filters that inspect and block malicious traffic directed at vulnerable devices;
    4) For encrypted traffic, we assume the edge security gateway can either decrypt the traffic (which is common for business ESGs such as Cisco Meraki~\cite{ciscomeraki}) using protocols like mcTLS~\cite{mctls}, or can infer malicious intent (which is also common for ESG Intelligence such as the Cisco Talos~\cite{merakilog4jint}) through analysis of metadata or unencrypted portions (e.g., packet headers, certificates).

\textbf{Baseline approach}:
To address the core challenges of scalability, robustness, and compatibility in exploit prevention, we can design a baseline framework integrating the strategic vantage point of Edge Security Gateways (ESGs) identified in Section~\ref{sec:relatedworks}, the generative capabilities of Large Language Models (LLMs), and the validation efficiency of Virtualized Network Function (VNF) to mitigate LLM hallucination. This streamlined pipeline operates as follows: upon disclosure of a 1-day/n-day vulnerability, an LLM-based generator dynamically crafts tailored filter rules leveraging Retrieval Augmented Generation (RAG), addressing scalability by eliminating manual rule creation and enabling rapid response within the first day. These rules then undergo rigorous validation through VNF testing against synthetic exploit traffic and benign traffic samples, which identifies and filters erroneous outputs before deployment. Finally, validated rules are propagated to ESGs, achieving compatibility by providing a unified enforcement layer that protects diverse connected devices across heterogeneous network environments.

\begin{figure}[t]
\vspace{-5pt}
\centering
\includegraphics[width=0.95\linewidth]{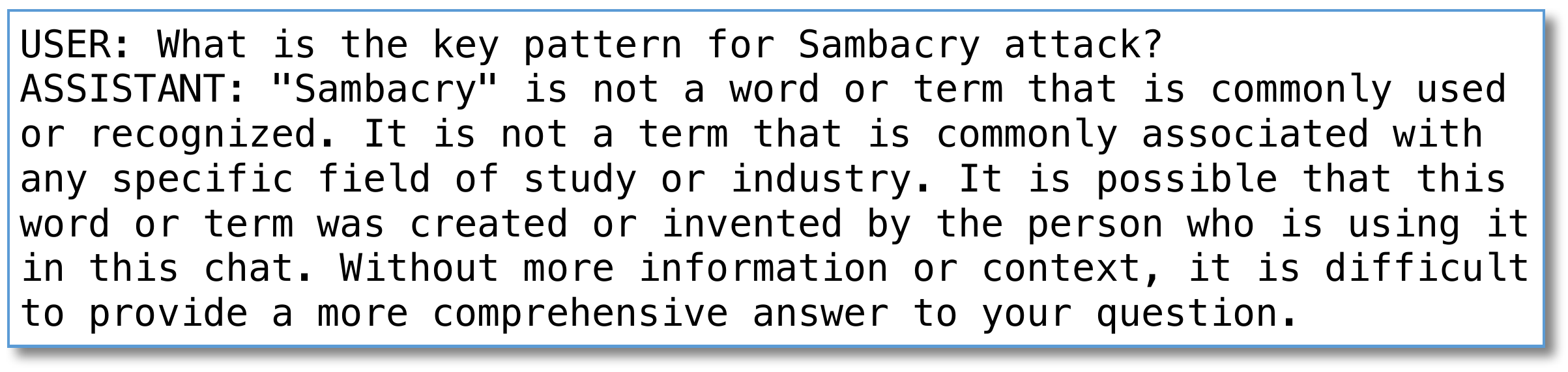}
\caption{Gemma-3-4B (May-2025) lacks expertise.}
\vspace{-5pt}
\label{fig:basicpatching_sambacry}
\end{figure}

\textbf{Challenges}:
While the baseline approach shows promise, several practical challenges hinder its implementation:

\begin{packeditemize}

\item \textbf{Limited Vulnerability-Fixing Expertise}:
Existing LLMs~\cite{mscopilot, chatgpt, llama, gemma3_12b} are trained on general-purpose datasets and lack specialized knowledge in niche domains like vulnerability remediation (Figure~\ref{fig:basicpatching_sambacry}). While techniques like Retrieval Augmented Generation (RAG) can incorporate vulnerability-related context into prompts, they fail to internalize domain-specific expertise. 
This limitation prevents LLMs from reliably inferring accurate filter fixes for novel \nday vulnerabilities


\Commentout{
Current LLM models~\cite{mscopilot, chatgpt, llama, vicuna, redpajama} are trained on generic dataset and lack the key domain-specific context for network-fixes.
For example, in Figure~\ref{fig:basicpatching_sambacry}, the Llama model~\cite{llama} (with 7B parameters) is prompted to generate key pattern for the SambaCry~\cite{sambacry} but do not understand the vulnerability context.
}


\item \textbf{Gaps Between Language and Network}:
Current LLMs are designed and optimized for natural language interactions, as exemplified by training paradigms like Reinforcement Learning from Human Feedback (RLHF). This creates a significant semantic and structural disconnect when applied to network security. Their effectiveness in processing raw network-layer data (e.g., packets) or generating precise protocol-specific rules remains unproven, limiting their practical utility for real-world network enforcement tasks.




\item \textbf{LLM Hallucination and Non-Determinism}: 
LLM outputs are inherently non-deterministic, often producing inconsistent or contradictory rules across repeated iterations. This instability introduces reliability risks, as identical inputs may generate divergent outputs—some of which could be erroneous or unsafe. Such unpredictability undermines trust in automated systems requiring consistent, repeatable results for critical tasks like vulnerability-fixing filtering.

    
\end{packeditemize}

\begin{figure}[t]
\centering
\includegraphics[width=0.5\textwidth]{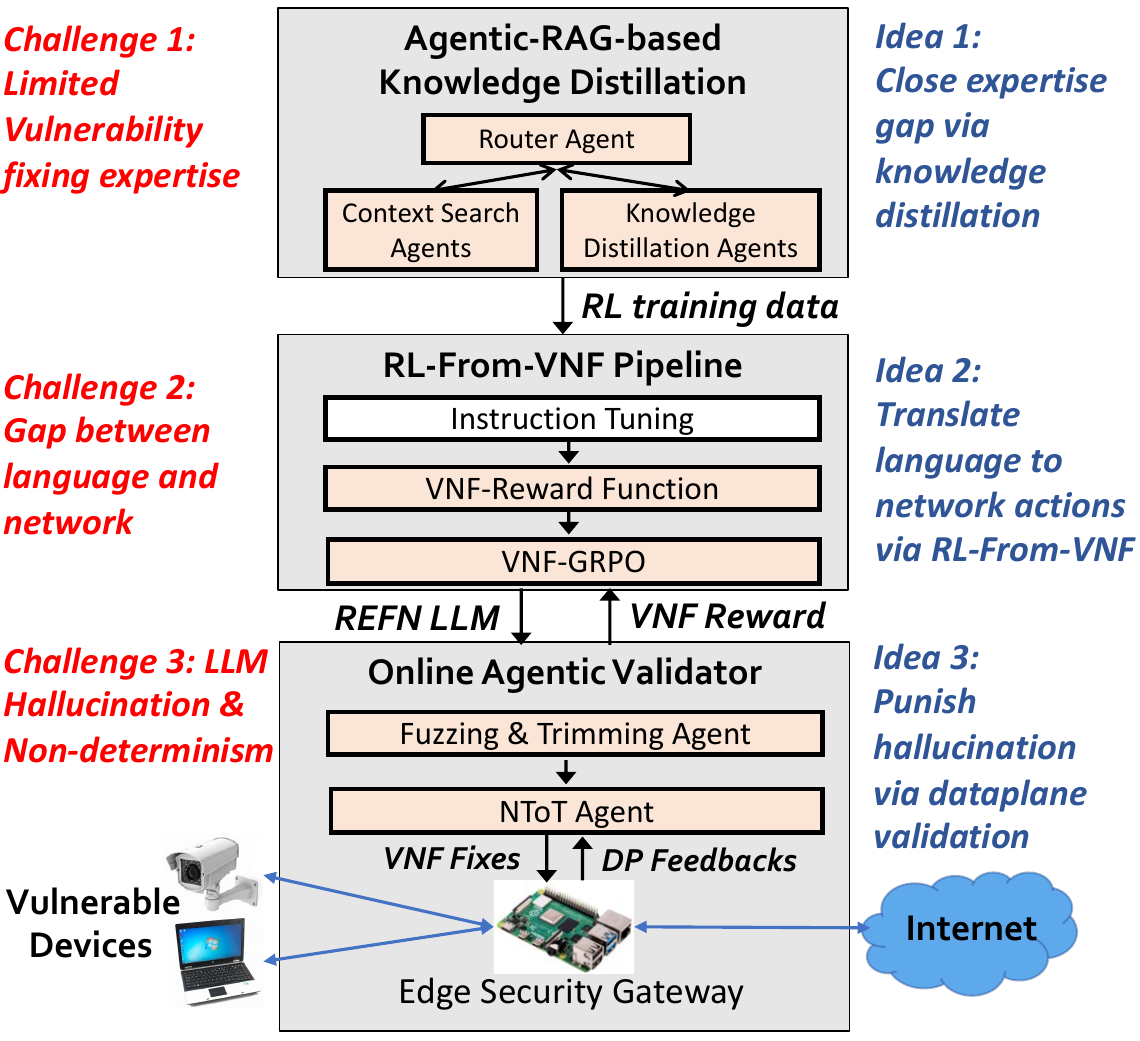}
\tightcaption{\Name's workflow.}
\vspace{-5pt}
\label{fig:overview_arch}
\end{figure}

\textbf{\Name's ideas}:
To address the above challenges, we propose \Name on top of the baseline approach, as shown in Figure~\ref{fig:overview_arch}.
\Name is built on a simple yet powerful premise: training LLMs into dynamic, network-aware vulnerability fixing engines that can deliver fixes on day one.
To achieve this, the system directly tackles the three core challenges through strategic design.
The first idea is to \textit{close the expertise gap via knowledge distillation}.
Instead of relying on generic LLM knowledge, \Name distills historical vulnerability fixes (e.g., past CVEs, patches) into the model during training. This equips the LLM with an implicit playbook of remediation strategies, enabling it to infer fixes for new 1-day vulnerabilities by recognizing patterns from past fixes—even if the new vulnerability differs superficially.
The second idea is to \textit{translate language to network actions via RL-from-VNF pipeline}.
\Name treats network filtering as a ``language'' the LLM must learn.
Through reinforcement fine-tuning (ReFT), the model is trained to enforce vulnerability-fixing filters on raw network packets (e.g., dropping malicious payloads) rather than generating text. Rewards are tied to real-world outcomes - blocking malicious packets while preserving legitimate packets.
The third strategy is to \textit{punish hallucination via dataplane validation}.
Every LLM generated filter is validated as Virtualized Network Function (VNF) with both benign and malicious traffic. Filters that fail to block attacks or disrupt legitimate flows are punished via a VNF-based online reward function, creating a feedback loop that force out the LLM hallucination in the training stage.

\textbf{\Name's workflow}:
As illustrated in Figure~\ref{fig:overview_arch}, REFN transforms raw 1-day/n-day vulnerability context into reliable security enforcements on Edge Security Gateways (ESGs) through three core components:

\textbf{Agentic-RAG-Based Knowledge Distillation}:
This pipeline integrates agent-based systems, Retrieval-Augmented Generation (RAG), and knowledge distillation to transfer vulnerability-fixing expertise from powerful-but-expensive LLMs to specialized models (efficient for training fixes for new vulnerabilities). 
This architecture features three autonomous agents: a Router Agent directing queries, Context Search Agents retrieving vulnerability context, and Knowledge Distillation Agents extracting structured inference examples (e.g., filter rules). 
This automated pipeline ensures RL-optimized knowledge transfer while eliminating manual expertise bottlenecks (detailed in Section~\ref{sec:rag}).




\textbf{RL-From-VNF Pipeline}:
Distilled knowledge is processed by the RL-From-VNF LLM training pipeline, 
which replaces human feedback with Virtualized Network Function (VNF) validation, which automatically generate rewards/penalties based on security enforcement outcomes. 
To improve training inefficiency, REFN presents the \textit{VNF-GRPO} algorithm that
integrates GRPO optimization, LoRA adapters and VNF. 
The pipeline’s \textit{VNF-Based Reward Function} evaluates each LLM-generated filter on both benign and malicious traffic samples, 
training a lightweight network-aware LLM capable of translating textual vulnerability contexts into practical vulnerability-fixing enforcements on the ESGs (detailed in Section~\ref{sec:rlfn}).


\textbf{Online Agentic Validator}:
To address the LLM hallucination and non-determinism challenge, 
REFN’s Online Agentic Validator will validate the filters and generate quantitative rewards for the RL-From-VNF pipeline. 
Unlike coarse-grained network validation (e.g., binary block/no-block decisions), 
this validator employs fine-grained VNF-specific logic and can fuzz on LLM-generated filters to reward ``near-correct'' outputs (key for RL training convergence). 
This approach provides granular feedback for Reinforcement Fine-Tuning (ReFT), balancing precision with training stability while reducing hallucinations. 
The design enables effective RL convergence and reliable exploit prevention (detailed in Section~\ref{sec:validation}).

\tightsection{Agentic-RAG-Based Knowledge Distillation}
\label{sec:rag}

\begin{figure}[t]
\centering
\includegraphics[width=0.9\linewidth]{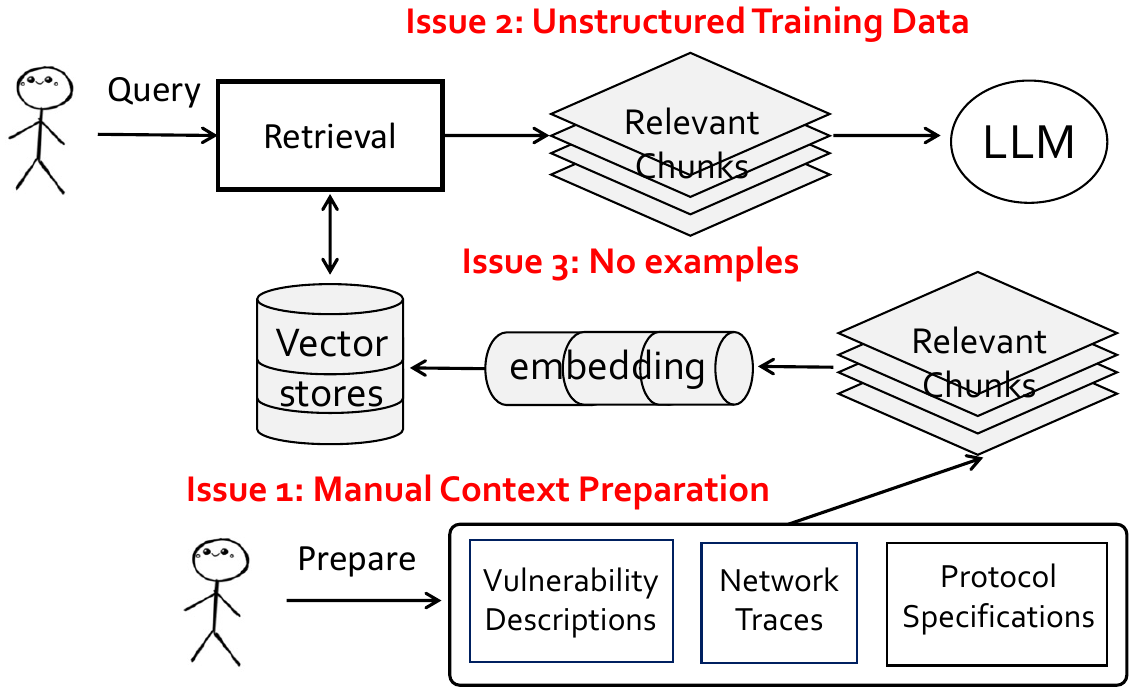}
\includegraphics[width=0.9\linewidth]{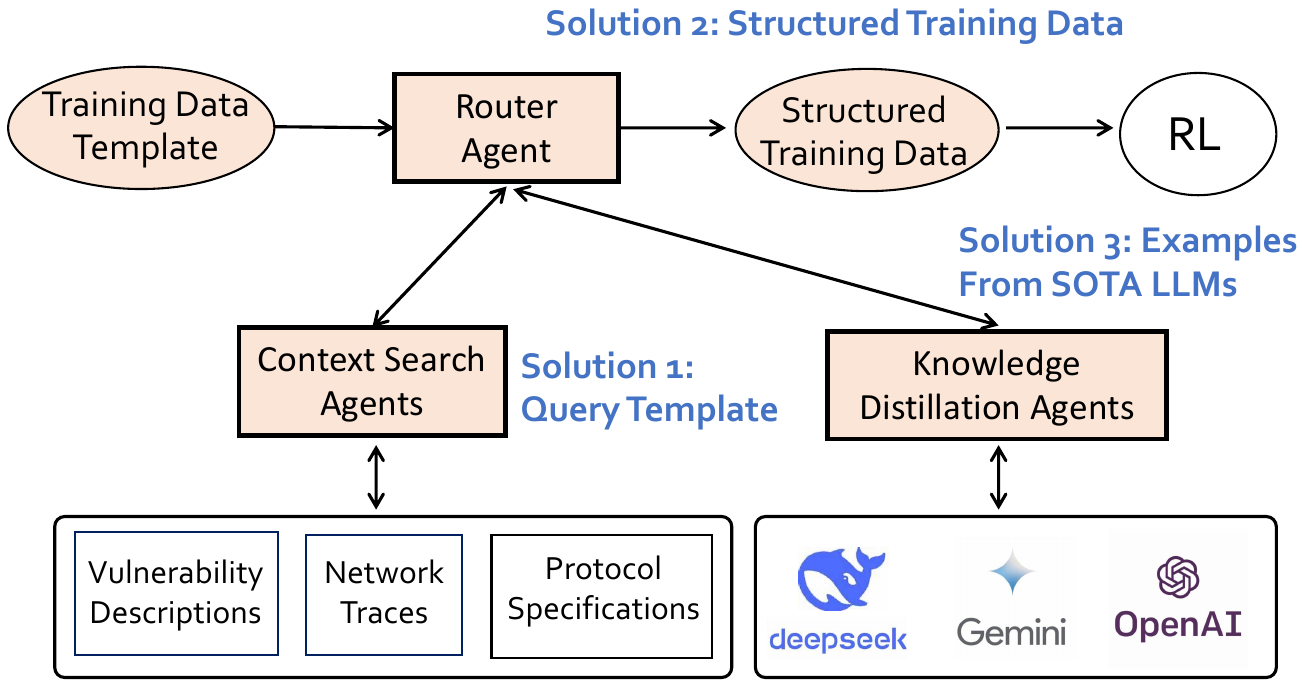}
\caption{Common RAG (top) vs REFN's Agentic-RAG-Based Knowledge Distillation (bottom).}
\vspace{-5pt}
\label{fig:refn_rag_agentrag}
\end{figure}

When a 1-day/n-day vulnerability emerges, the Reinforcement Learning (RL) process requires curated training data, which currently is prepared manually with the common Retrieval-Augmented Generation (RAG) (Figure~\ref{fig:refn_rag_agentrag} top).
This conventional approach suffers from three critical limitations: 
labor-intensive preparation of vulnerability contexts (descriptions, network traces, protocol specifications), unstructured blended data that compromises RL effectiveness, and inability to distill knowledge from SOTA LLMs (DeepSeek-R1, Gemini-2.5, GPT-4o).
To overcome these constraints, we introduce the Agentic-RAG Knowledge Distillation pipeline (Figure~\ref{fig:refn_rag_agentrag} bottom), which integrates agent-based systems, RAG, and knowledge distillation to transfer capabilities from expensive LLMs (e.g., DeepSeek-R1-671B) to efficient-to-train specialized models.
The pipeline employs three autonomous components: a \textit{Router Agent} that dynamically directs queries, \textit{Context Search Agents} that retrieve vulnerability intelligence, and \textit{Knowledge Distillation Agents} that extract structured inference samples from SOTA LLMs—automating knowledge transfer while ensuring RL-optimized, structured outputs.

\begin{figure}[t]
\centering
\includegraphics[width=0.9\linewidth]{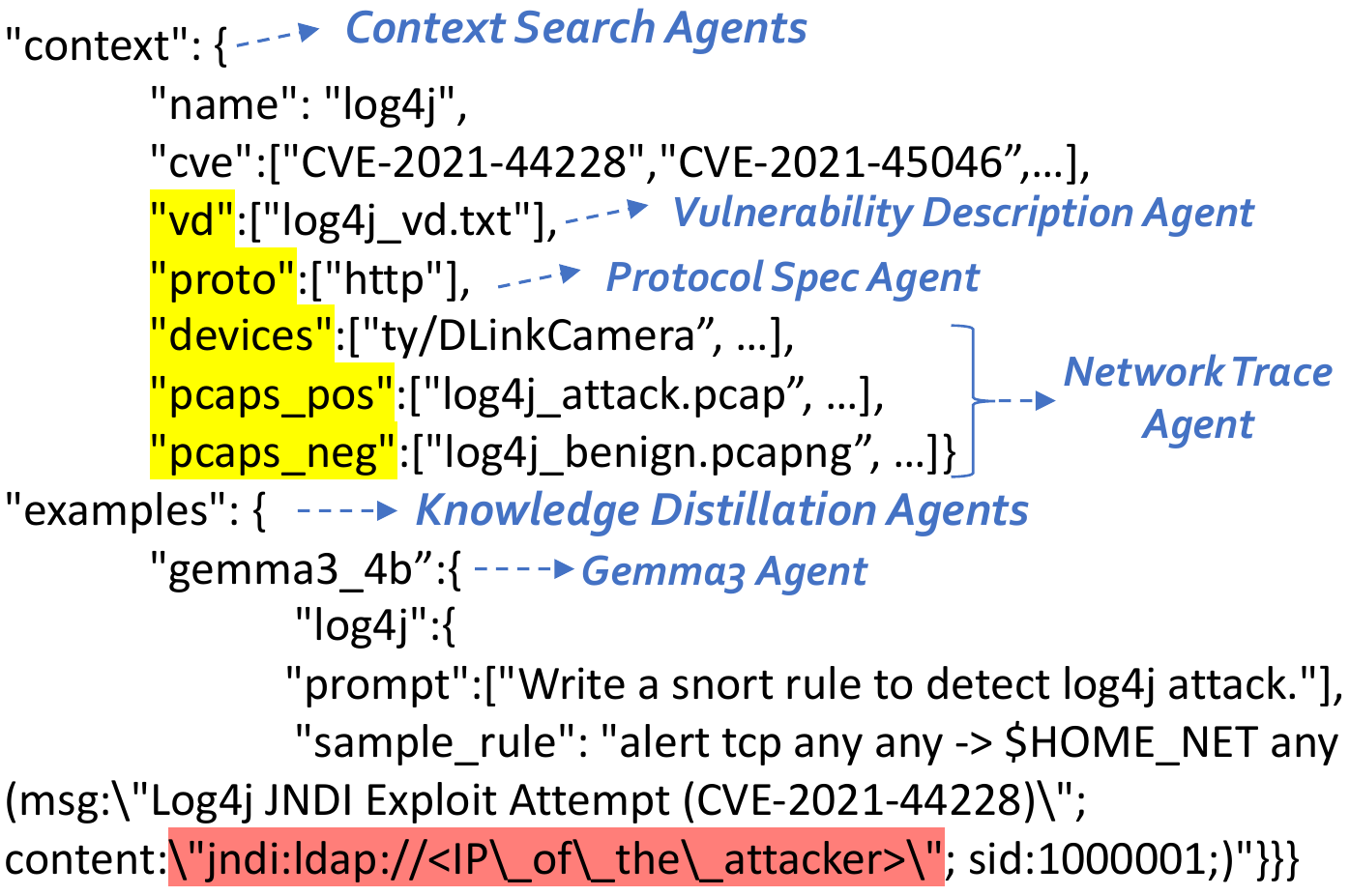}
\caption{Training data template.}
\vspace{-10pt}
\label{fig:refn_rag_datatemplate}
\end{figure}

\begin{figure}[t]
\centering
\includegraphics[width=0.9\linewidth]{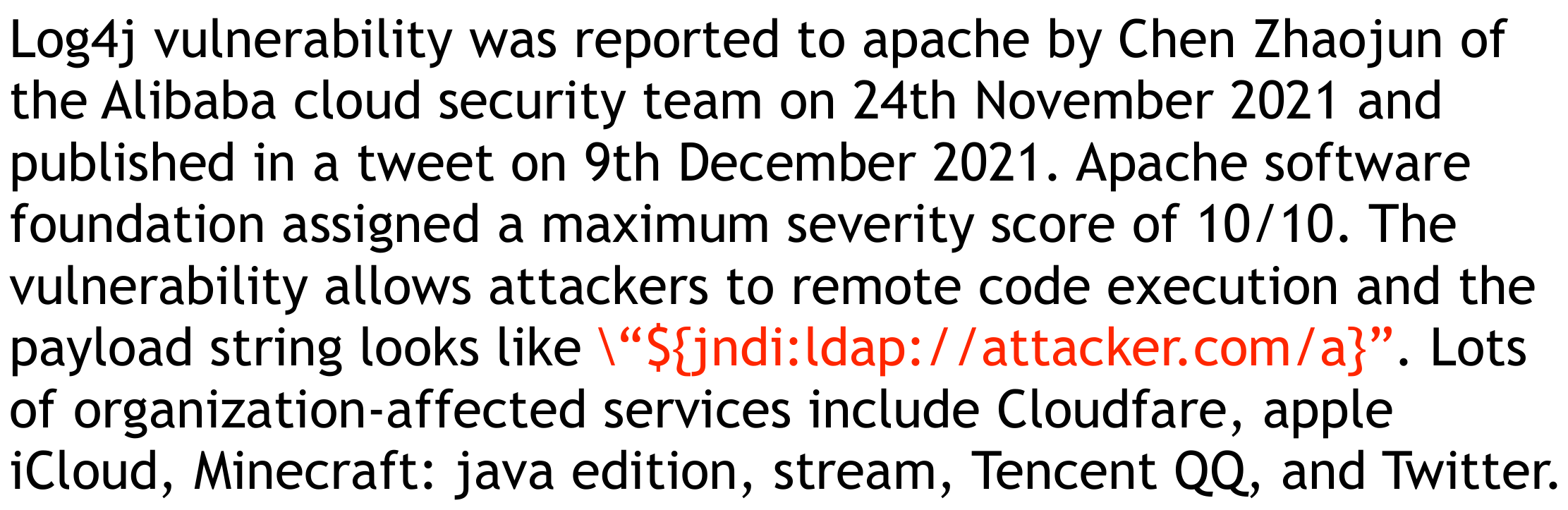}
\tightcaption{Segment of Log4j~\cite{log4jvul} vulnerability description.}
\vspace{-5pt}
\label{fig:basicpatching_vd_log4j}
\end{figure}

\Commentout{
\begin{figure}[t]
\centering
\includegraphics[width=0.9\linewidth]{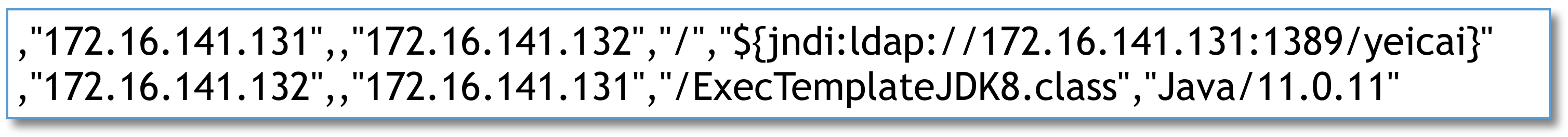}
\tightcaption{Segment of Log4j~\cite{log4jvul} exploit traffic log.}
\vspace{-5pt}
\label{fig:basicpatching_nt_log4j}
\end{figure}
}


\tightsubsection{Router Agent}

When a \nday vulnerability first emerges, 
REFN initiates training data preparation by sending a structured prompt to the Router Agent (RA).
In this prompt,
REFN presents a well-designed RL-training data template (Figure~\ref{fig:refn_rag_datatemplate}) that captures key RL-training structures for exploit prevention.
Upon receiving the training-data-preparation prompt, the agent decomposes data preparation tasks
based on the template and delegates them to specialized context search agents and knowledge distillation agents. 
Three context search agents operate in parallel: 
(1) The vulnerability description agent extracts vulnerability details from CVE databases and security advisories~\cite{nvd}, such as the Log4j vulnerability description in Figure~\ref{fig:basicpatching_vd_log4j}; 
(2) The protocol specification agent identifies relevant network protocols and their specifications for trace parsing; 
(3) The network trace agent retrieves packet captures (from public repositories~\cite{netresec, iot23, dvrhack}), populating positive/negative traffic examples (``pcap\_pos/pcap\_neg'') and device context. 
Concurrently, the knowledge distillation agent queries SOTA LLMs for vulnerability-fixing filter examples. 
While these examples may contain inaccuracies (e.g., Snort rules content ``jndi:ldap://IP\_of\_the\_attacker'' with correct prefixes like ``jndi:ldap'' but erroneous continuations like ``IP\_of\_the\_attacker''), 
they provide valuable structural patterns. This enriched data significantly enhances Supervised Fine-Tuning (SFT), ultimately improving output quality throughout the RL pipeline.

\tightsubsection{Context Search Agents}

The context search agent prepares critical vulnerability context, including descriptions (vd), network traces (pcaps), and protocol specifications (proto). A naive non-agent approach relies on conventional Retrieval-Augmented Generation (RAG) with in-context learning (ICL): first crawls vulnerability descriptions from sources like NVD~\cite{nvd}, network traces from repositories~\cite{netresec, iot23, dvrhack}, and protocol specifications from IETF~\cite{httpieft}, then chunks this heterogeneous data indiscriminately for ICL prompting. However, the unstructured blending of data types significantly compromises retrieval efficacy~\cite{min2022icl}.

To address this limitation, REFN introduces a \textit{vulnerability-to-trace pair-ranking} mechanism that explicitly correlates vulnerability descriptions $vc$ with network traces $nc$, enhancing context preparation efficacy. Our approach segments the vulnerability description context $vc$ into discrete sentence-level contexts $x^{vc}$ and the network trace context $nc$ into packet-level contexts $x^{nc}$. It then generates sentence-packet context pairs $(x^{vc}, x^{nc})$ and algorithmically selects the subset of pairs exhibiting the highest semantic correlation $\{(x^{vc}, x^{nc})\}$.
Based on previous research~\cite{min2022icl},
suppose $x$ is the testcase, $y$ is the ground truth for the testcase, $C$ is the space of all ground truth, and there are $k$ labeled examples $(x_1, \widehat{y_1}),..., (x_k, \widehat{y_k})$,
then the LLM output is:
$argmax_{y \in C} P(y | x_1, \widehat{y_1},..., x_k, \widehat{y_k}, x)$.
Consider each sentence-packet context pair $(x^{vc}, x^{nc})$ as a demonstration, the LLM output can be approximated by:
\begin{equation}
argmax_{y \in C} P(y | [(x^{vc}_1, x^{nc}_1),..., (x^{vc}_k, x^{nc}_k)]_{top\_k}, x) ,
\end{equation}
where $[(x^{vc}_1, x^{nc}_1),..., (x^{vc}_k, x^{nc}_k)]_{top\_k}$ are the top $k$ most correlated sentence-packet context pairs.
Such approximation is feasible because $nc \in C$, i.e., the network context $nc$ augmented from the vulnerable device's traces shares the same ground truth space $C$ with the desired output $y$.
In other words, the LLM output $y$ will be close to vulnerability descriptions $vc$'s real network traffic pattern.

To implement the sentence-packet context pair ranking efficiently, \Name provides a \textit{join labeling} operation:
\begin{equation}
[(x^{vc}_1, x^{nc}_1),..., (x^{vc}_k, x^{nc}_k)]_{top\_k} = vc \Join_{con} nc,
\end{equation}
where $con$ is $d(x^{vc}, x^{nc})<top\_k(\{d(x^{vc}, x^{nc}\})$ \footnote{$k=10$ unless specified otherwise.}.
More specifically, $d(x^{vc}, x^{nc})$ is the distance metrics between sentence context $x^{vc}$ and the packet context $x^{nc}$. We leverages FastText~\cite{fasttext} metrics for efficient distance calculation.

\tightsubsection{Knowledge Distillation Agents}

The Knowledge Distillation (KD) Agents extract and structure actionable insights from SOTA LLMs to prepare data for the Supervised Fine-Tuning (SFT) stage of RL training. Leveraging transfer learning principles, these agents adapt general knowledge from expensive Pre-trained Language Models (PLMs) to the specific downstream task of vulnerability remediation through three key mechanisms:

\textit{Knowledge Inheritance Mechanism}:
PLMs acquire intrinsic linguistic patterns (syntax, semantics, knowledge representation) through self-supervised pretraining on large corpora. The distillation framework inherits this knowledge as initialization, avoiding inefficient learning from scratch.
More specifically, given a list of known vulnerabilities $\{v\}$, a PLM $M$ and a filter generation prompt $p$, the KD agents will generate a list of filters $\{f\} = M(\{v\}, p)$ and prepare the knowledge as $(\{v\}, p, \{f\})$ for the SFT process.

\textit{Essence of Task Adaptation}:
By minimizing the vulnerability-fixing-specific loss function on the structured training data, the model parameters in the SFT process are adjusted to adapt to the filter generation domain:
    \begin{equation}
        \theta_{\text{SFT}} = \arg\min_{\theta} \sum_{(x,y) \in \mathcal{D}} \mathcal{L}\left( f(x; \theta), y \right)
        \label{eq:sft-optimization}
    \end{equation}
where $\theta$ represents the model parameters, $\mathcal{D}$ denotes the labeled dataset, $\mathcal{L}$ is the loss function (e.g., cross-entropy), and $f(\cdot)$ is the model's predictive output.

\textit{Catastrophic Forgetting Control}:
In SFT, a low learning rate strategy (typically 1-10\% of pretraining rates) combined with early stopping preserves foundational inference knowledge (e.g., generating filter rule) while accommodating vulnerability-fixing requirements (e.g., correct rule format).

Integrated into REFN's \textit{Instruction Tuning} stage (Section~\ref{sec:rlfn}), this distillation framework enables structured extraction of filtering rule patterns from SOTA LLMs, ensuring effective knowledge transfer to specialized security models.

\tightsection{RL-From-VNF Pipeline}
\label{sec:rlfn}

This section introduces the RL-From-VNF (RFV) pipeline, a novel framework that eliminates human feedback from the reinforcement learning loop. We first contrast RFV against state-of-the-art approaches (RLHF, DPO) to establish its unique network-driven paradigm. Next, we detail the pipeline's operational workflow and present its VNF-GRPO optimization algorithm. Finally, we analyze the core VNF reward function, demonstrating how network-generated feedback replaces human evaluations to autonomously drive RL training.


\tightsubsection{Comparison with RLHF and DPO}

\begin{figure}[t]
\centering
\includegraphics[width=0.95\linewidth]{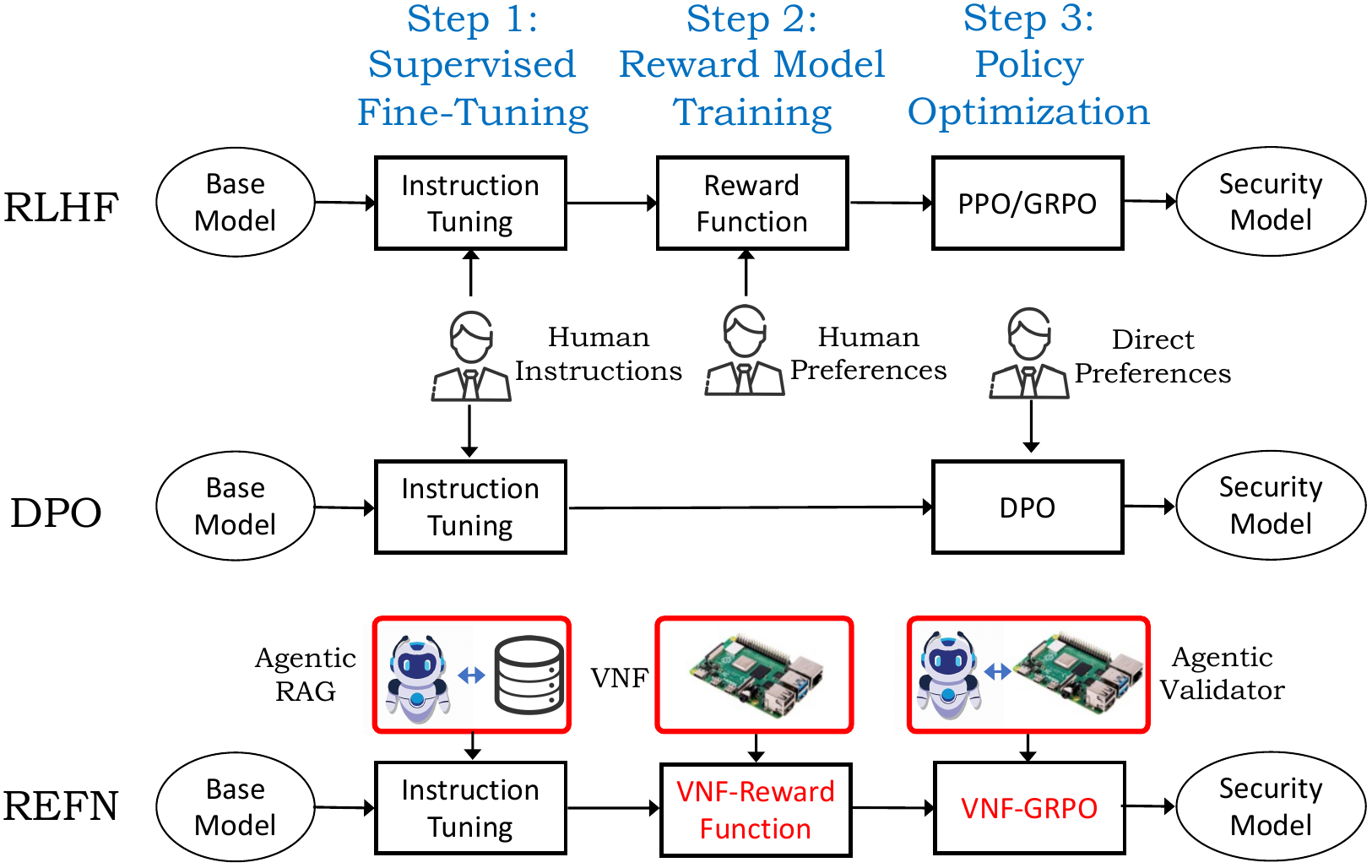}
\tightcaption{RL-From-VNF Pipeline.}
\vspace{-5pt}
\label{fig:refn_pipeline}
\end{figure}

As shown in Figure~\ref{fig:refn_pipeline}, we compare the RL-From-VNF pipeline with
two SOTA RL methodologies: Reinforcement Learning from Human Feedback (RLHF) and Direct Preference Optimization (DPO).

\textbf{\textit{RLHF:}} This approach
includes three key stages.
First, supervised fine-tuning (SFT) leverages human-generated instruction data to align a pre-trained model's initial outputs with security objectives, such as filter rule generation. 
Second, reward modeling directly incorporates human feedback, where annotators rank or rate model responses to train a preference predictor that quantifies qualitative judgments into reward signals. Finally, policy optimization employs algorithms like PPO~\cite{ppo} or GRPO~\cite{shao2024deepseekmath} to iteratively refine the model against this reward predictor, embedding human preferences throughout the optimization cycle. 

\textbf{\textit{DPO:}} This approach
streamlines preference integration while maintaining human dependency. Similar to RLHF, DPO begins with SFT using human-crafted instruction-response pairs for initial task alignment. However, its policy optimization phase diverges significantly: annotators rank output pairs, and the DPO algorithm directly optimizes for preferred responses using a contrastive loss function that compares favorable against dispreferred outputs. This approach eliminates the explicit reward modeling step required in RLHF, instead embedding human feedback directly within the optimization process while reducing implementation complexity.

\textbf{\textit{Reliance on human feedback:}}
The reliance on human feedback in both RLHF and DPO workflows makes them ill-suited for fixing \nday vulnerabilities, which demand real-time, automated responses. 
First, human feedback introduces latency: RLHF requires iterative human annotation to train reward models, while DPO depends on pre-collected human preference datasets. These steps create delays incompatible with the time-sensitive nature of \nday exploits, where threats must be neutralized within hours or minutes of discovery. 
Second, human feedback loops struggle to scale with the sheer volume and diversity of network traffic—annotators cannot feasibly label every potential intrusion pattern in dynamic network environments. 
Third, human biases or inconsistencies in labeling could inadvertently prioritize non-critical alerts or miss adversarial evasion tactics, undermining precision. 



\textbf{\textit{RL-From-VNF idea:}}
We propose the RL-From-VNF (RFV) pipeline—a paradigm where Virtualized Network Functions (VNFs) replace human feedback in the reinforcement learning loop. Unlike traditional RL that relies on handcrafted reward functions (e.g., ``maximize throughput''), RFV leverages security VNFs deployed on edge gateways to autonomously validate vulnerability-fixing filters. These VNFs generate real-time reward/penalty signals based on concrete security outcomes, such as successful malicious traffic blocking without disrupting benign flows. This approach eliminates human subjectivity while harnessing domain-specific validation logic to guide RL optimization.
By integrating VNF validation logic (e.g., firewall checks) as the reward generator, RFV eliminates manual reward engineering—a significant departure from conventional RL systems.
This architecture directly enables zero-touch networking principles by establishing closed-loop automation: security policies self-optimize through continuous feedback from live network enforcement, requiring minimal human intervention while maintaining context-aware precision.

\tightsubsection{VNF-GRPO Algorithm}

\begin{algorithm}[t]
\caption{VNF-GRPO Algorithm}
\label{alg:nfv_grpo}
\begin{algorithmic}[1]
\STATE \textbf{Input:} 
Initial policy $ \pi_\theta $ with parameters $\theta$, \\
\textcolor{blue}{batch of trajectories \( \mathcal{D} = \{(x, e, p, y_{\emptyset})\} \),}\\ 
\textcolor{blue}{NFV reward function \( \mathcal{R} \),}\\
learning rate \( \beta \), 
regularization factor \( \lambda \), 
clipping factor \( \epsilon \), 
number of group members \( N \)

\STATE \textbf{Output:} Updated policy $ \pi_\theta' $ with parameters \( \theta' \)

\STATE // Initialize policy parameters \\
$\theta \gets \theta^{(0)}$

\FOR{each member $i$ from $1$ to $N$}
    \STATE \textcolor{blue}{// Collect trajectory $ \mathcal{D}_i $ from member $i$}
    \STATE \textcolor{blue}{$ VNF_i \gets \text{VNF\_GENERATOR}(x,e,i) $}
    \STATE \textcolor{blue}{$ \mathcal{D}_i \gets \text{VNF\_VALIDATOR}(p \rightarrow VNF_i ) $}

    \STATE \textcolor{blue}{// Compute VNF reward $ r_i $}
    \STATE \textcolor{blue}{$r_i \gets \mathcal{R} (\mathcal{D}_i)$}
    \STATE // Compute group-relative reward $ R_i $
    \STATE $ R_i = \frac{r_i}{\sum_{j=1}^{N} r_j} $

    \STATE \textcolor{blue}{// Compute the advantage function $A_i$}
    \STATE \textcolor{blue}{$A_i \gets \text{VNF\_ADVANTAGE}(R_i)$}
\ENDFOR

\FOR{each member $i$ from $1$ to $N$}
    \STATE // Compute the surrogate objective function: \\
    $\mathcal{L}_i(\theta) = SURROGATE\_OBJ(\pi_\theta, A, \epsilon)$
    \STATE // Compute Gradient Ascent and Update the policy:\\
    $ \theta_i = \theta_i + \beta \cdot \nabla_{\theta} \mathcal{L}_i(\theta) $
\ENDFOR

\FOR{each agent \( i \) from 1 to \( N \)}
    \STATE Apply regularization to stabilize the policy update: \\
    $\mathcal{L}_{\text{regularized}}(\theta) = \mathcal{L}(\theta) - \lambda \cdot \text{penalty}(\theta)$
    \STATE Update the policy parameters \( \theta_i \) with regularization: \\
    $\theta_i = \theta_i - \lambda \cdot \nabla_{\theta} \text{penalty}(\theta)$
\ENDFOR

\STATE \textbf{Return:} Updated policy parameters \( \theta' \)
\end{algorithmic}
\end{algorithm}


Algorithm~\ref{alg:nfv_grpo} presents the VNF-GRPO algorithm and its differences (in blue) with the basic GRPO algorithm~\cite{shao2024deepseekmath}.

\textbf{Input:} 
The standard GRPO inputs include 
initial policy $ \pi_\theta $ with parameters $\theta$, 
learning rate $\beta$, 
regularization factor $\lambda$, 
clipping factor $\epsilon$, 
number of group members $N$.
The two distinguished inputs include batch of trajectories $\mathcal{D}$ and VNF reward function $\mathcal{R}$.
The basic GRPO batch of trajectories $\mathcal{D} = \{(x, e, y)\}$
is tuples of question $x$, CoT $e$ and answer $y$.
However, since the dependency of Human Feedback is removed,
\Name's batch of trajectories' answer $y_{\emptyset}$ is an empty set.
The VNF feedbacks relies on the processing result on the online packets $p$.
Therefore, \Name's batch of trajectories can be denoted as $\mathcal{D} = \{(x, e, p, y_{\emptyset})\}$.


\textbf{Update trajectories, rewards and advantages from VNF:} 
After the policy parameters are initiated (Line 3),
the trajectories $\mathcal{D}_i$ will be updated for each member $i$ (Line 5-7). 
The VNF\_GENERATOR will generate a virtualized function $VNF_i$
for each member $i$ (Line 6).
The VNF\_GENERATOR will update the trajectories based on 
the packets processing result of the VNF $p \rightarrow VNF_i$ (Line 7).
The VNF reward function $\mathcal{R}$ will compute the reward $r_i$ based on the trajectories (Line 9), which is converted into the group-relative reward $R_i$ (Line 11).
The VNF\_ADVANTAGE function will compute the advantage function $A_i$ for member $i$ based on the the group-relative reward $R_i$,
which estimates the difference between the expected reward for an action and the average reward.
Traditionally, the human feedbacks decide which actions are preferable given a certain state.
In \Name, it is replaced by the VNF feedbacks via the VNF\_ADVANTAGE function.


\textbf{Update Policy with Group Relative Consideration:}
This part includes the step to Compute Surrogate Objective
and the Gradient Ascent on Surrogate, which follows the basic GRPO.
The $SURROGATE\_OBJ(\pi_\theta, A, \epsilon)$ function is:

\begin{equation}
\mathbb{E}\left[ \min\left( \frac{\pi_{\theta}(a|s)}{\pi_{\theta_{\text{old}}}(a|s)} A(s, a),
\text{CLIP} (\pi_\theta, s, a, \epsilon) A(s, a) \right) \right],
\end{equation}
where $\pi_\theta$ is the policy with parameter $\theta$, $A$ is the advantage function, $a$ is the action, $s$ is the state,
$\epsilon$ is the clipping factor.
The CLIP function is:

\begin{equation}
\text{CLIP} (\pi_\theta, s, a, \epsilon) = 
\text{clip}\left( \frac{\pi_{\theta}(a|s)}{\pi_{\theta_{\text{old}}}(a|s)}, 1-\epsilon, 1+\epsilon \right),
\end{equation}
After that, the Gradient Ascent on Surrogate is computed and the policy is updated (Line 17).

\textbf{Apply Regularization for Stability:}
To ensure stability, a regularization term is applied based on the relative group performance (Line 20-21). This is a penalty term that discourages large changes in the policy parameters.

\tightsubsection{VNF Reward Function}
A naive reward function implementation would assign binary rewards per benign/malicious sample (e.g., per pcap file). However, such coarse-grained signals prove insufficient for effective reinforcement learning guidance~\cite{shao2024deepseekmath}, necessitating finer-grained alternatives: per network flow, per Application Data Unit (ADU), or per packet.
The network packets are dynamic under various network configurations (e.g., switch MTU) and is not suitable as the basic unit for rewarding.
The network flow are hard to assemble and hard to compare, which is key for assigning reward.
Consequently, we choose the ADU, which is a consecutive sequence of packets from one device to another, 
leveraging its established adoption in network penetration testing methodologies~\cite{buzz} and inherent stability across network conditions.



Consider a benign sample (pcap) as a sequence of ADUs $B$ and a malicious sample (pcap) as a sequence of ADUs $M$.
A simple way to assign reward is to find out how much ADUs are blocked in $B$ (FP) and how much ADUs are allowed in $M$ (FN).
However, this approach fundamentally misrepresents malicious sample $M$, where most ADUs constitute benign background traffic and only a small portion of ADUs in $M$ is related to attack payloads. 
Penalizing these benign ADUs within $M$ introduces contradictory signals that destabilize reinforcement learning. To resolve this distortion, we apply a differentiation operation to isolating and excluding benign ADUs in $M$ before reward calculation:
\begin{equation}
M - B = \{ m: m \in M \&\& m \notin PairRank(M,B)\}
\end{equation}
where $PairRank(M,B)$ pairs the ADUs in $M$ and $B$, rank them by their similarity and output the same ADUs in sequence $M$ and sequence $B$.

Based on these observations, we define the reward function:
\begin{equation}
\mathcal{R} = 2pc/(p+c) 
\end{equation}
which is basically the F1-Score reward that address the unbalanced data (malicious data is much fewer than benign data), where the precision reward $p = \frac{TP(M - B)}{TP(M - B)+FP(B)}$,
and the recall reward $c = \frac{TP(M - B)}{TP(M - B)+FN(M - B)}$. 
Note that in the above reward calculation, instead of tracking all the TP/TN/FP/FN in benign ADU sequence $B$ and malicious ADU sequence $M$,
we optimized to only track TP/FN in $M - B$ and FP in $B$.
Given that $M$ is much less than $B$, such optimization would greatly reduce the validation cost while providing effective reward to guide the RL process for the unbalanced security data.






\tightsection{Online Agentic Validator}
\label{sec:validation}

\begin{figure}[t]
\centering
\includegraphics[width=0.9\linewidth]{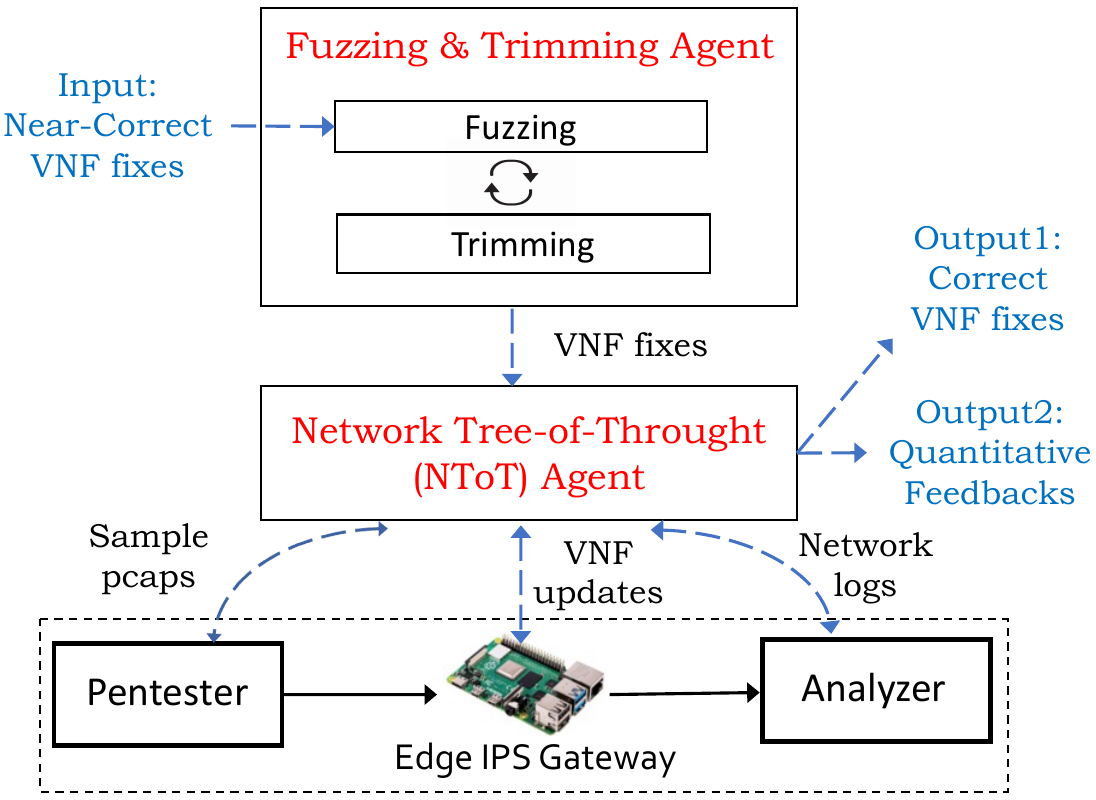}
\tightcaption{\Name's validation workflow.}
\vspace{-5pt}
\label{fig:validation_workflow}
\end{figure}

To address LLM hallucination issues, \Name introduces an Online Agentic Validator, as shown in Figure~\ref{fig:validation_workflow}. 
This solution targets two critical RL failure modes: 
1) \textit{Near-correct outputs}: generated \netpatch that are near-correct but contain subtle discrepancies; 
2) \textit{Feedback granularity gap}: require quantitative diagnostic feedback to drive reinforcement learning beyond binary validation.
\Name's validator resolves these challenges through two integrated components: 1) A \textbf{Fuzzing \& Trimming Agent} that bridges the correctness gap by iteratively refining near-correct \netpatch into fully valid fixes via constraint-guided mutation;
2) A \textbf{Network Tree-of-Thought (N-ToT) Agent} that analyzes live penetration testing results to generate fine-grained metrics, providing the structured quantitative feedback required for RL optimization. 
This dual approach not only rectifies hallucination-induced inaccuracies but also establishes a closed-loop online system for continuous RL to prevent emerging \nday exploits.








\begin{figure}[t]
\centering
\includegraphics[width=0.95\linewidth]{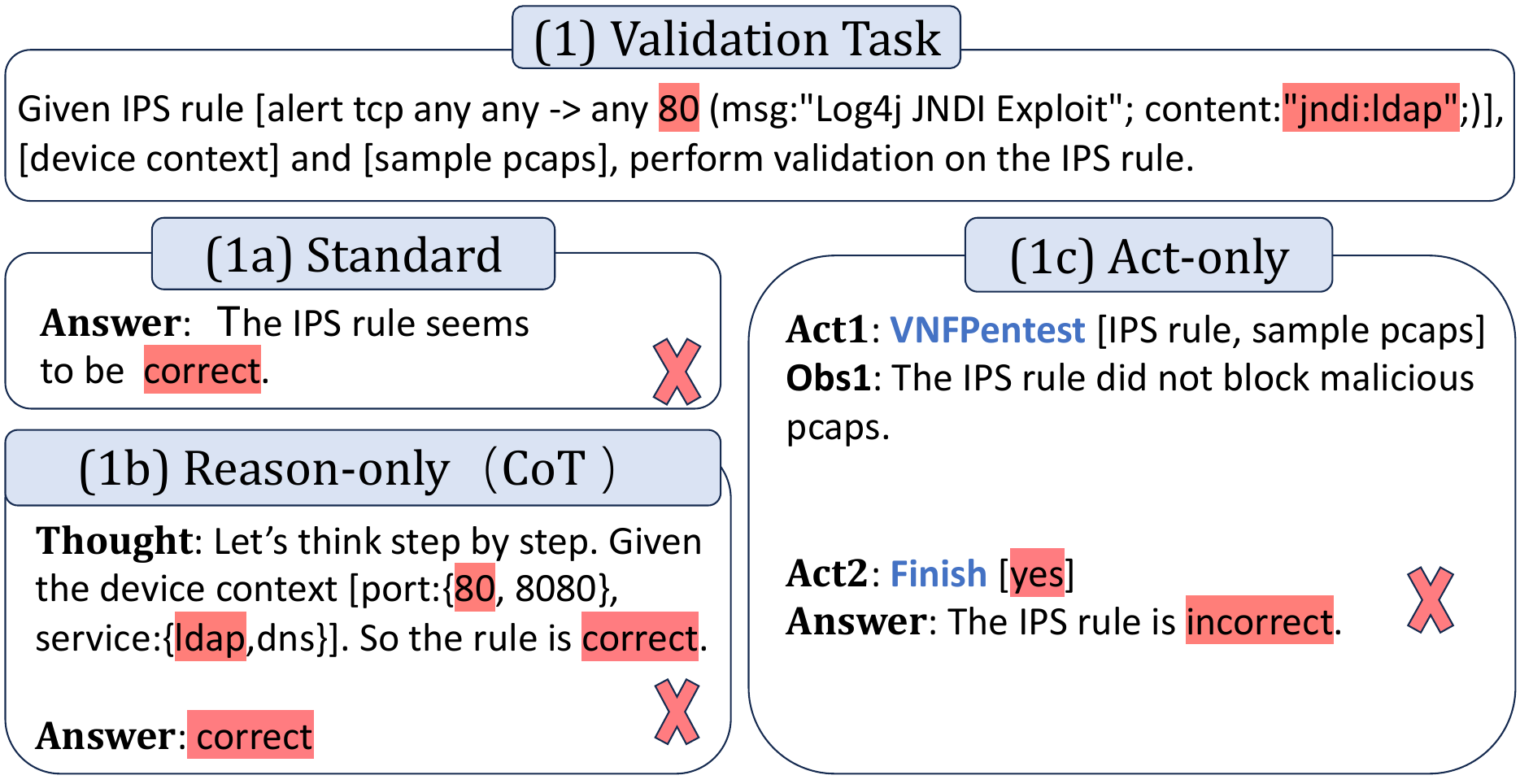}
\includegraphics[width=0.95\linewidth]{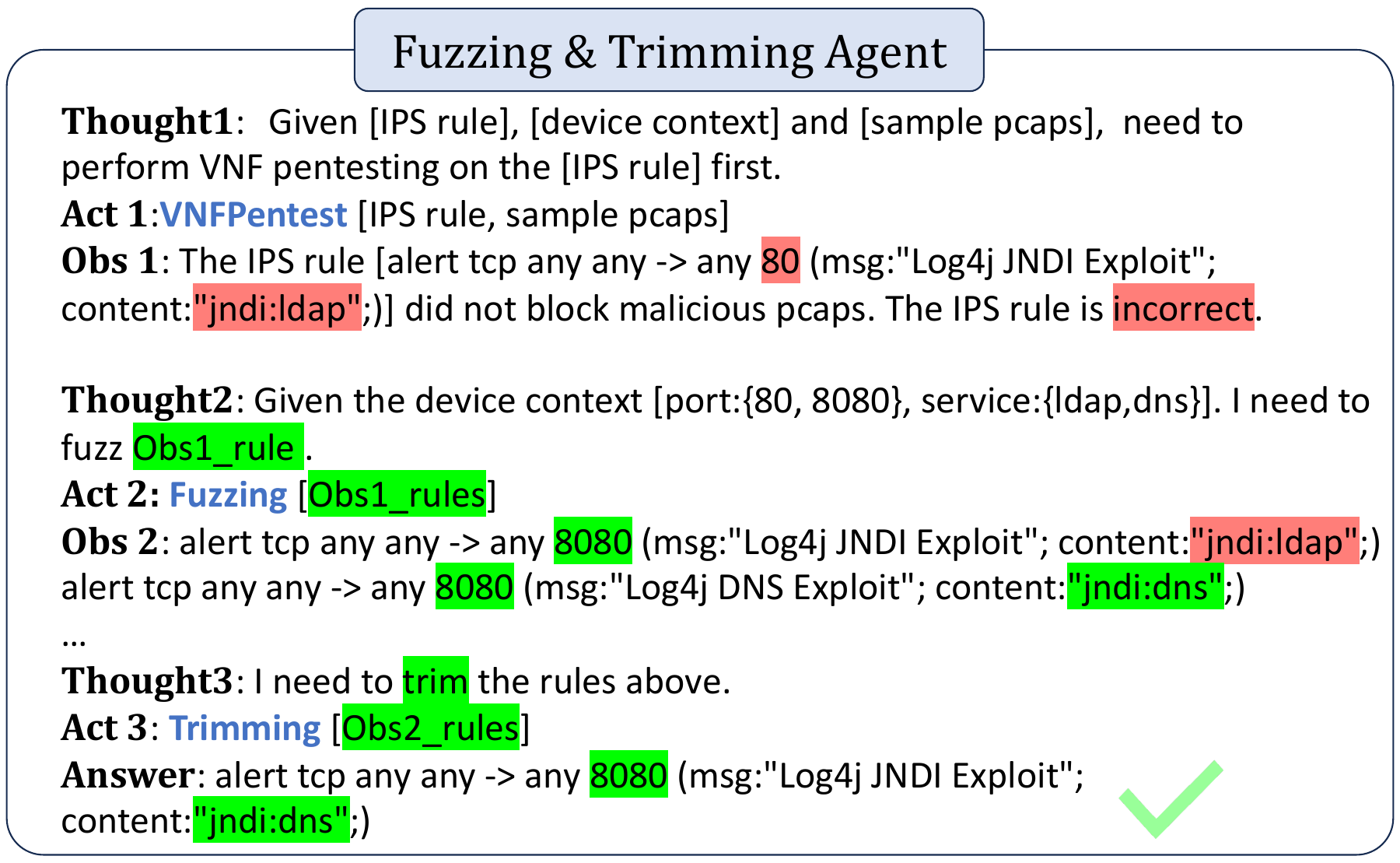}
\tightcaption{\Name's fuzzing \& trimming agent.}
\vspace{-5pt}
\label{fig:refn_agent_fuzz}
\end{figure}

\tightsubsection{Fuzzing \& Trimming Agent}

The first challenge is handling \textit{near-correct outputs}. Current LLM methodologies—including standard LLMs, Reason-only Agents, and Act-only Agents—are ill-suited for this purpose. Consider the validation task in Figure~\ref{fig:refn_agent_fuzz}: given a near-correct filter rule (with a mismatched port 80 and partially incorrect content ``jndi:ldap''), device context, and sample pcaps, validate the rule's effectiveness. 
A standard LLM merely validates the rule's format, falsely concluding it is correct. A Reason-only Agent performs limited context matching (e.g., checking ports 80/8080 and services ldap/dns) but also fails to detect the flaw. Only an Act-only Agent, by performing a \textit{VNFPentest} (enforcing the rule on benign and malicious pcaps), identifies that malicious traffic bypasses the rule. 
However, the Act-only Agent only provides a binary pass/fail result that lacks the ability to refine the near-correct rule itself.

To address this limitation, \Name developed a fuzzing \& trimming agent by extending the ReAct framework~\cite{yao2023react}, specifically designed to refine near-correct \netpatch into fully valid fixes. 
The agent employs a two-phase approach: first performing fuzzing to explore solution space through controlled randomness (``random walk''), followed by trimming to eliminate defective branches (e.g., malformed rules) and optimize search efficiency. 
As Figure~\ref{fig:refn_agent_fuzz} demonstrates, the agent begins by executing a \textit{VNFPentest} (Act 1) 
and observe that the filter rule cannot block the malicious pcaps.
However, it did not stop but continue to perform Fuzzing (Act 2)
on the \textit{near-correct} rule, generate variants via random walk towards the correct solution.
During this process, invalid branches are systematically pruned. 
Ultimately, the agent concluded with the right filter rule (with port 8080 and content ``jndi:dns'').


\begin{algorithm}[t]
\small
    \caption{NToT-BFS} \label{algorithm:ntot_bfs}
    \begin{algorithmic}[1]
        \REQUIRE dataplane ADUs $ADU_{D}$, validation agent $AG_{v}$, network protocol spec $netsp$, middlebox spec $boxspec$

        \ENSURE middlebox action $C_m \in (BLOCK, ALLOW, ALERT)$

        \Comment{Create decision tree $DT$}
        \STATE $DT \gets AG_{v} (netsp, boxspec)$
        
        \Comment Create thought generator $G()$
        \STATE $G() \gets AG_{v} (DT, netsp, boxspec)$

        \Comment Create states evaluator $V()$
        \STATE $V() \gets AG_{v} (DT, netsp, boxspec)$

        \Comment Perform NToT-BFS
        \STATE $S_0 \gets Init(DT)$
        
        \STATE $\textbf{for}$ $adu \in ADU_{D}$ $\textbf{do}$
        \begin{ALC@g}
        \STATE $g = G(DT, S_t)$ 
        \STATE $S_{t+1} = V(adu, DT, g)$
        \STATE $t \gets t+1$
        \STATE $C_m \gets S_{t+1}$
        \end{ALC@g}
        \STATE $\textbf{end}$ $\textbf{for}$
        \STATE return $C_m$
    \end{algorithmic}
\end{algorithm}

\tightsubsection{Network Tree-of-Thought (NToT) Agent}
The second challenge is the \textit{feedback granularity gap}, as quantitative feedbacks beyond binary result is required for RL process.
REFN addresses this through its Network Tree-of-Thought (N-ToT) Agent, which analyzes live penetration testing results to generate structured quantitative feedback for RL optimization. As illustrated in Figure~\ref{fig:validation_workflow}, the workflow begins when the NToT agent receives VNF fixes from the Fuzzing \& Trimming agent and deploys them to the Edge IPS Gateway via VNF updates. The agent then coordinates penetration testing by: (1) configuring benign and malicious test cases for the Pentester, and (2) deploying expected dataplane actions to the Analyzer. During execution, the Pentester conducts tests while the Analyzer compares observed traffic against expected actions, returning network logs to the NToT agent. This enables the agent to validate VNF fixes and derive quantitative RL feedback—a complex task requiring inference of middlebox enforcement from dynamic dataplane traffic. For example, during Log4j vulnerability testing, the agent must precisely calculate the blocking rate for malicious payloads (e.g., "jndi:dns") while measuring false positive rates on benign traffic.
While LLM agents typically employ Chain-of-Thought (CoT) for multi-stage inference~\cite{langchain}, this method proves inadequate for middlebox analysis due to two fundamental limitations: locally, CoT fails to explore alternative reasoning paths within a thought process; globally, it lacks state planning mechanisms to evaluate different continuation options across the inference trajectory.

To address these issues, \Name provide the NToT (Network-Tree-of-Thought) inference mechanism, as shown in Algorithm~\ref{algorithm:ntot_bfs}.
The key idea of NToT is to build a decision-tree thought structure $DT$ based on network and middlebox specifications (Line 1).
Then, the agent will map the dataplane ADUs to the states of decision-tree $DT$ and infer the middlebox actions (Line 4-11) based on the following components:

\textit{Thought generator}: NToT provide a thought generator $G()$ to generate thought for each state $S_t$ in decision-tree $DT$ (Line 6).
The thought generator is created based on the decision tree $DT$, the network protocol specifications $netspec$ and the middlebox specifications $boxspec$ (Line 2).

\textit{State evaluator}: NToT provide a state evaluator $V()$
to evaluation multiple continuations at each state $S_t$ based on current dataplane ADU $adu$, decision-tree $DT$ and current thought $g$ (Line 7).
The state evaluator $V()$ is created based on the decision tree $DT$, the network protocol specifications $netspec$ and the middlebox specifications $boxspec$ (Line 3).

\Name's NToT stores the decision tree structure and allows for the addition, deletion, or modification of nodes and edges. It also supports the inclusion of new protocols. \Name enables the easy analyze the live penetration testing results to generate fine-grained metrics and provide structured quantitative feedback (based on the VNF-Reward Function in Section~\ref{sec:rlfn}) required for RL optimization.

\section{Dataset and Implementation}
\label{sec:implementation}

\begin{table}[t]
\scriptsize
\begin{tabular}{ p{3.0cm} | p{4.5cm}  }

\hline
Exploit Family & Vulnerability Examples \\

\hline
Log4j & CVE-2021-44228, CVE-2021-45046  \\

\hline
SambaCry& CVE-2017-7494  \\

\hline
Mirai & CVE-2020-5902  \\

\hline
Modbus Injection & CVE-2022-1068  \\

\hline
Spooky SSL & CVE-2022-3602, CVE-2022-3786 \\

\hline
Eternal Blue & CVE-2017-0143, CVE-2017-0148  \\

\hline
CryptoWall & CVE-2015-5560, CVE-2015-5122  \\

\hline
AlexaEavesDropping & CVE-2023-33248  \\

\hline
CiscoRouterCmp & CVE-2017-3881  \\

\hline
CiscoRouterHttp & CVE-2024-20393  \\

\hline
HuelightBlackout & CVE-2020-6007  \\

\hline
PlexDVRExp & CVE-2020-5741 \\

\hline
ToriiBot & CVE-2023-1389 \\

\hline
Virut & CVE-2014-4114 \\

\hline
Yakesmalware & CVE-2014-0224 \\





\hline
Conficker & CVE-2008-4250  \\









\hline
Gh0st RAT & CVE-2018-8174,
CVE-2012-0507  \\



\hline
Locky Ransomware & CVE-2012-0507, CVE-2015-5122  \\



\hline
Pony & CVE-2017-11882  \\


\hline
Bladabindi & CVE-2017-8759  \\





\hline
WannaCry & CVE-2017-0144, CVE-2017-0145 \\

\hline
Trojan.Valyria & CVE-2017-11882  \\



\hline

\end{tabular}
\caption{The 22 families of \nday exploits.}
\label{tab:patch_cases}
\end{table}

\textbf{Dataset}:
We present the first dataset that enables RL-training of LLMs to prevent 1-day/n-day exploits.
The dataset is generated by using \Name's Agentic-RAG-Based Knowledge Distillation module to gather four key parts of data:
1) \textit{vulnerability descriptions}: details from CVE databases and security advisories~\cite{nvd}, such as the Log4j example in Figure~\ref{fig:basicpatching_vd_log4j}; 
(2) \textit{protocol specifications}: network protocols and their specifications~\cite{httpieft} for trace parsing; 
(3) \textit{network traces}: packet captures (pcaps) from online repositories including NETRESEC~\cite{netresec}, IoT-23~\cite{iot23} and IoT Sentinel~\cite{iotsentinel}, populating positive/negative traffic examples (``pcap pos/neg'') and device context.
The 22 families of 1-day/nday
exploits (Table~\ref{tab:patch_cases}) and benign samples from 65 types of devices.
As shown in Figure~\ref{fig:refn_rag_datatemplate}, 
for each family of exploit (e.g., Log4j), there is a list of vulnerabilities (``cve'' field), a vulnerability description text (``vd'' field), corresponding devices (``devices'' field), malicious pcaps (pcaps\_pos) and benign pcaps (``pcaps\_neg'' field).

\textbf{Implementation}:
We implemented \Name with 6K LoC on two desktop servers (one for RL training and one for testbed) and a edge security gateway. 
Each server is equipped with an \texttt{NVIDIA RTX 4090 GPU} (24GB VRAM), \texttt{Intel Platinum 8352 CPU} (36 cores), 32GB RAM, and 16TB HDD.
The edge security gateway is implemented on a \texttt{Raspberry Pi 4B} running \texttt{Snort 2.9.8.0}.
\Name's base LLM model for RL training and agents is \texttt{Gemma 3-4B}. 
The agents integrates \texttt{ReAct} framework and is deployed using \texttt{Ollama}~\cite{ollama}.
The RAG is implemented using \texttt{LangChain}~\cite{langchain} (chunk size = 500 and chunk overlap = 10). The vector store is using \texttt{FAISS}~\cite{faiss}.





\tightsection{Evaluation}
\label{sec:evaluation}

\begin{table*}[t]
\centering
\small
\caption{Overall effectiveness comparison of \Name and alternative approaches.}
\label{tab:evaluation_effectiveness_all}
\begin{tabular}{ p{5.5cm} | p{1.5cm} | p{1.5cm} | p{1.5cm} | p{1.5cm} }
\hline
Approaches & FPR & FNR & Accuracy & F1-Score \\  
\hline
\manualpatch & 0.068 & 0.819 & 0.556 & \textbf{0.290} \\ 
\hline
\softwarepatch & 0.034 & 0.910 & 0.528 & 0.161 \\ 
\hline
\gmlpatch & 0.017 & 0.881 & 0.551 & 0.209 \\ 
\hline
\gllmpatch & 0.033 & 0.904 & 0.531 & 0.170 \\ 
\hline
\networkfiltering & 0.032 & 0.893 & 0.537 & 0.188 \\ 
\hline
\gmlfiltering & 0.144 & \textbf{0.630} & 0.819 & 0.238 \\ 
\hline
\gllmfiltering & \textbf{0} & 1 & \textbf{0.925} & 0 \\
\hline
\Name & \textbf{0.003} & \textbf{0.071} & \textbf{0.992} & \textbf{0.945} \\ 
\hline
\Name's improvement & - & $\geq 88.7\%$ & $\geq 21.1\%$ & $\geq 225.9\%$ \\ 
\hline
\end{tabular}
\vspace{-15pt}
\end{table*}


In this part, we evaluate \Name and show that:

\begin{packeditemize}
    \item \Name is \textit{effective}, with $\geq 21.1\%$ accuracy improvement and $\geq 225.9\%$ F1-Score improvement than alternatives.

    \item \Name is \textit{efficient} - Mean-Time-To-Patch (MTTP) is 3.65h (95.4\% improvement);
    the fix installation delay (iDelay) is at second-scale (10X reduction) compared with alternatives.

    \item \Name is \textit{scalable} - the Batched Training Time (BTT) for 22 \nday vulnerabilities is less than 0.5 day; can easily be applied to 10000 vulnerable devices (around 300 offices) with 1.5 hours of Accumulative Downtime (ADT) in total.
    
\end{packeditemize}

\tightsubsection{Experiment Setting}

\textbf{Basic setting}:
We established a penetration testbed (Figure~\ref{fig:evaluation_topo} in Appendix~\ref{sec:appendix_evaluation_testbed}) to evaluate \Name and alternative approaches (Table~\ref{tab:evaluation_effectiveness_all}). This testbed incorporates attack launchpads and user interfaces connected to target devices via the edge security gateway. 
%
The testbed integrates both physical devices and virtualized devices (using QEMU~\cite{qemu}) to host the 22 families of \nday exploits (Table~\ref{tab:patch_cases} in Section~\ref{sec:implementation}). 
The testbed leverages virtualization to enables scalable hosting of expensive/hard-to-replicate embedded devices (e.g., smart grid transformers).
A dedicated server hosting the \Name framework will generate the
\netpatch and deploy them on the edge security gateway for enforcement.

\textbf{Alternative approaches}: 
As shown in Table~\ref{tab:evaluation_effectiveness_all}, we evaluated seven categories of alternative vulnerability fixing approaches for comparison:
1) \manualpatch: we conducted an IRB-approved study with 10 security admins patching devices using official documentation;
2) \softwarepatch: we tested 9 business-grade solutions (detailed in Table~\ref{tab:patch_management} in Appendix~\ref{sec:appendix_softwarepatch_list})~\cite{patchmanager, chocolatey, avira, ninite, patchupdater, sumo, heimdal, npackd, ruckzuck};
3) \gmlpatch: we evaluated typical ML-based patching approaches including GraphSPD~\cite{graphspd}, RNNPatch~\cite{patchrnn}, PAVUDI~\cite{pavudi} and SPI~\cite{spi};
4) \gllmpatch: we evaluated the patch generation capability of 
ChatGPT-4o~\cite{chatgpt}, DeepSeek-R1~\cite{deepseekr1} and Gemma3-12B~\cite{gemma3_12b} (2025-July version);
5) \networkfiltering: we conducted an IRB-approved study with 10 security admins creating Snort 2.9.8.0~\cite{snort} rules;
6) \gmlfiltering: we evaluated ML-based network filtering approaches including Kitsune~\cite{kitsune} and ODDS~\cite{odds};
7) \gllmfiltering: we evaluated the network filtering rule generation capability of ChatGPT-4o~\cite{chatgpt}, DeepSeek-R1~\cite{deepseekr1} and Gemma3-12B~\cite{gemma3_12b} (2025-July version).
More details about the IRB-approved study are presented in Section~\ref{sec:irb}.
To have fair comparison, for the seven categories of alternative approaches, we synthesized the best results across methods to represent the category's performance.




\begin{figure}[t]
\centering
\includegraphics[width=0.5\textwidth]{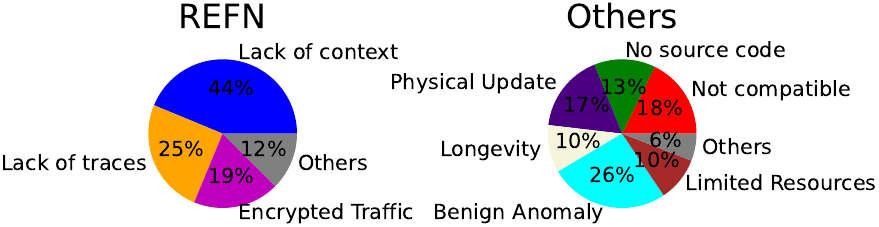}
\tightcaption{Pie graph analysis of invalid fixes.}
\vspace{-5pt}
\label{fig:evaluation_effectiveness_pie}
\end{figure}

\tightsubsection{Effectiveness}

\textbf{\textit{Overall effectiveness}}:
We evaluate the effectiveness of \Name and alternative approaches
over 22 families of \nday exploits and benign samples from 65 types of devices
(detailed in \Name's dataset Section~\ref{sec:implementation})
on four metrics:
1) \textit{FPR} (False Positive Rate), defined as $\mathit{\frac{FP}{FP + TN}}$;
2) \textit{FNR} (False Negative Rate), defined as $\mathit{\frac{FN}{FN + TP}}$;
3) \textit{Accuracy}, defined as $\mathit{\frac{TP + TN}{P + N}}$;
4) \textit{F1-Score}, defined as $\mathit{\frac{2*Precision*Recall}{Precision + Recall}}$.
We choose these four metrics because the \nday exploit scenario is highly biased (benign samples $>$ malicious samples)~\cite{bhatt2014operational}.
Therefore, relying solely on Accuracy is dangerously misleading, as a model can achieve high scores by always predicting benign.
FPR reveals how often legitimate activities are wrongly flagged (causing operational disruption), while FNR measures the failure to detect actual attacks (leading to undetected breaches). 
The F1-score provides a crucial balanced metric over both FPs and FNs, offering a more complete picture of model effectiveness. 

Table~\ref{tab:evaluation_effectiveness_all} presents the performance of \Name and alternative approaches using FPR, FNR, Accuracy, and F1-Score. While the \gllmfiltering approach achieves the lowest FPR (0) and second-highest Accuracy (0.819), its critical flaw is revealed by its FNR of 1.0 – it misclassifies all malicious samples as benign. Our analysis attributes this failure to the severe LLM hallucination during rule generation, producing seemingly correct but ultimately flawed filtering rules (exemplified in Figure~\ref{fig:refn_rag_datatemplate}). \Name effectively addresses this limitation; leveraging RL training with negative rewards for detected false positives, it achieves the second-lowest FPR (0.003). Crucially, \Name achieves the lowest FNR (0.071), representing an 88.7\% reduction compared to the next best (generic ML-based at 0.627). Although the \gmlfiltering approach detects some \nday exploits, its reliance on network traffic anomalies results in the highest FPR (0.144), causing significant disruption to benign traffic. For F1-Score, \manualpatch achieves the second-highest (0.290), but its manual nature hinders scalability. Overall, \Name demonstrates superior effectiveness, delivering $\geq 21.1\%$ higher Accuracy and $\geq 225.9\%$ higher F1-Score than the best alternative approaches.

\textbf{\textit{Root Cause Analysis}}:
We investigated the reasons behind ineffective fixes for both \Name and alternative approaches, as depicted in Figure~\ref{fig:evaluation_effectiveness_pie}.
For alternative approaches, the key factors are: 
1) benign anomaly (26\%), key factor for high FPs in ML-based filtering;
2) not compatible (18\%), key factor for FNs in patching; 
3) need physical update such as serial cable (17\%); 
4) no source code available (13\%); 
5) limited resources (10\%); 
6) vendor's longevity issues (10\%).
For \Name, the key factors are: 1) lack of contexts (44\%); 2) lack of traces (25\%); 3) encrypted traffic (19\%).
Different from other approaches, \Name is not impacted by the compatibility issue.
Notably, the inference mechanism can alleviate the impact of encrypted traffic (only 19\%).
The lack of contexts issue in \Name is caused by the accessibility of the training data including vulnerability descriptions and traces, and can be alleviated by crowd sourcing in the future.

\begin{table}[t]
\scriptsize
\caption{Time-To-Patch (TTP) for \Name.}
\label{tab:evaluation_ttp}
\begin{tabular}{ p{2.5cm}| p{1.0cm}| p{1.0cm} | p{1.0cm} | p{1.0cm} }
\hline
Time-To-Patch & Min & Max & P90 & Mean \\  
\hline
\manualpatch & 3d 19.34h & 7d 20.85h & 5d 17.58h  & 5d 9.62h \\ 
\hline
patch software & 3d 5.92h & 7d 3.47h & 5d 11.19h & 4d 22.29h \\ 
\hline
\networkfiltering & 1d 21.23h & 5d 4.79h & 4d 18.76h & 3d 6.61h \\ 
\hline
\Name & 2.42h & 5.29h & 4.57h & 3.65h  \\ 
\hline
\Name improvement & $94.7\%$ & $95.8\%$ & $96.0\%$ & $95.4\%$ \\ 
\hline
\end{tabular}
\end{table}

\tightsubsection{Efficiency}

Evaluating the efficiency of fixing \nday vulnerabilities requires addressing two critical questions:

\begin{packeditemize}
    \item \textit{Can vulnerability fixing outpace exploitation?}
    Specifically, given the \textbf{Time-To-Patch (TTP)} – defined as the duration from vulnerability exposure to the deployment of a protective fix – does the system achieve a TTP consistently less than one day? This threshold represents the minimum exploitation window defined for \nday vulnerabilities.
    
    \item \textit{Does the fix installation minimize operational disruption?}
    Specifically, given the \textbf{Installation-Delay (iDelay)} – defined as the operational downtime imposed on a normal device during the fix deployment process – the iDelay should be low and ensuring minimal disruption to benign devices.
\end{packeditemize}


    

We evaluate the efficiency of \Name in terms of Time-To-Patch (TTP) and Installation-Delay (iDelay), and compare it with \manualpatch, \softwarepatch, \networkfiltering (in the IRB study, 10 admins manually create patches or network filters).
We are unable to measure the TTP and iDelay for common ML and LLM patching/filtering approaches, because these approaches only provide trained models and their training time and deployment time are not disclosed (e.g., DeepSeek-R1 or the auto-encoder training in Kitsune~\cite{kitsune}).
However, the TTP for common ML and LLM patching/filtering approaches is estimated to be \textit{a few days to several week} based on online reports~\cite{deepseektrainingtime}, exceeding the critical 1-day threshold for outpacing the \nday exploit.



\textbf{Time-To-Patch (TTP)}:
Table~\ref{tab:evaluation_ttp} presents the Min, Max, P90 (90th percentile), and mean Time-To-Patch (TTP) metrics for REFN. 
As shown in Table~\ref{tab:evaluation_ttp}, the Mean-Time-To-Patch (MTTP) for manual patching and patch management software is at least 4 days 22 hours. This high duration stems from the inherent complexity of host-based patching. For instance, remediating a Log4j vulnerability on an Apache server requires crafting JVM patches, testing compatibility and effectiveness, and redeploying both the JVM and Apache server – a costly procedure. 
Similarly, manual network filtering also incurs a significant MTTP of at least 3 days and 6 hours due to the intricate, error-prone process of crafting rules from vulnerability descriptions, parsing exploit packets, and thoroughly testing and adjusting the filters against both malicious and benign traffic.
In contrast, REFN achieves a dramatically lower MTTP of just 3.65 hours. This represents a 95.4\% improvement over the next fastest method (manual network filtering at 3 days 6.61 hours). Crucially, REFN's fixing speed is significantly faster than the critical 1-day threshold associated with 1-day/n-day exploits. 
This efficiency is attributed to REFN's key components: its Genetic-RAG-Based Knowledge Distillation for rapid data collection, the RL-from-NFV pipeline for efficient model training, and the Online Agentic Validator for swift validation and adjustment.

\begin{figure}[t]
\centering
\includegraphics[width=0.48\textwidth]{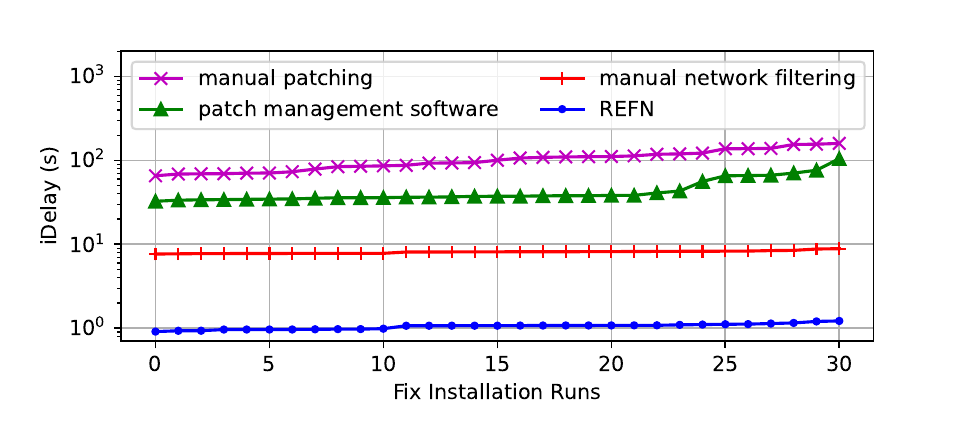}
\tightcaption{Installation-Delay (iDelay).}
\label{fig:evaluation_efficiency_downtime}
\vspace{-5pt}
\end{figure}

\textbf{Installation-Delay (iDelay)}:
Figure~\ref{fig:evaluation_efficiency_downtime} presents the iDelay of manual patching, patch management software, manual network filtering, and REFN. iDelay measures the operational downtime imposed on a normal device during vulnerability fix deployment. To ensure a comprehensive evaluation, we conducted multiple runs (31) per approach over   fixable vulnerabilities for all approaches, and sorted the results in ascending order. REFN achieves a significantly lower iDelay than alternatives, consistently requiring only around $1$ second per deployment. 
In contrast, manual patching and patch management software incur delays of several minutes, primarily due to mandatory device/software restarts in host-based patching.
Manual network filtering exhibits an iDelay of around 10 seconds, attributed to the need for administrator intervention and IPS hot restarts.
This translates to REFN delivering at least a 10x reduction in iDelay compared to the next fastest feasible approach (manual network filtering) and a remarkable around 80x reduction compared to patching methods. This efficiency stems from REFN's RL-supported and validated Virtual Network Function (VNF) deployment, leveraging agile middlebox update techniques~\cite{khalid2016paving} for near-instantaneous updates and minimum
disruption.

\begin{figure}[t]
\centering
\includegraphics[width=0.43\textwidth]{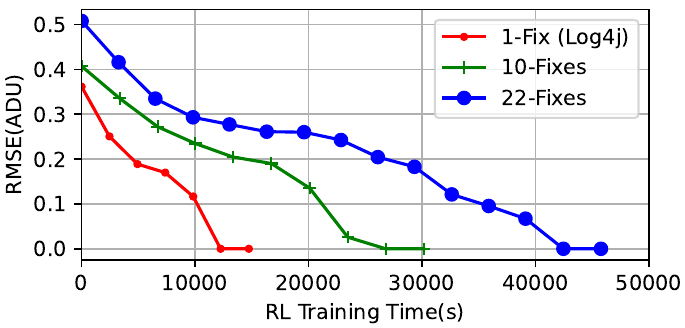}
\tightcaption{Batched Training Time (BTT).}
\vspace{-5pt}
\label{fig:evaluation_trainingtime}
\end{figure}

\begin{table}[t]
\scriptsize
\caption{Accumulative Downtime (ADT).}
\label{tab:evaluation_downtime}
\begin{tabular}{ p{2.5cm}| p{1.0cm}| p{1.0cm} | p{1.0cm} | p{1.0cm} }
\hline
Devices & 10 & 100 & 1000 & 10000 \\  
\hline
\manualpatch & 1136s & 12182s & - & - \\ 
\hline
patch software & 689s & 6608s & 65521s & - \\ 
\hline
\networkfiltering & 82s & 1038s & - & -  \\ 
\hline
\Name & 6s & 56s & 575s & 5486s  \\ 
\hline
\end{tabular}
\vspace{-5pt}
\end{table}

\tightsubsection{Scalability}

There are two key questions that needs to be answered in evaluating the scalability of fixing 1-day/n-day vulnerabilities:

\begin{packeditemize}
    \item \textit{How do the training time increases when the number of vulnerabilities scales up? Can the training be batched?}
    We define \textbf{Batched Training Time (BTT)} as REFN's time cost for training multiple vulnerabilities at once.
    
    \item \textit{What is the overall operation disruption for fix installation when the number of devices scales up?}
    We define \textbf{Accumulative Downtime (ADT)} as the total time of all devices' normal function being disrupted by the fixing process.
\end{packeditemize}

\textbf{Batched Training Time (BTT)}:
As shown in We measured how \Name's batched training time (BTT) shift from training 1-fix (Log4j) to training 10-fixes, to training 22-fixes.
The x axis is the overall training time at each iteration.
An iteration is the process between the training start to the time the result is validated.
The y axis is the RMSE (Root Mean Square Error) of the result when validated in each iteration.
The RMSE is calculated based on correctly matched Application Data Units (ADUs) during the validation process.
In Figure~\ref{fig:evaluation_trainingtime}, the 1-Fix's (red line, Log4j) training process took 5 iterations to complete. 
The total training time was 3.40 hours ($12250$s).
The batched 10-fixes (green line line) took 8 iterations to complete.
The total training time was 7.45 hours ($26807$s).
The batched 22-fixes (green line line) took 14 iterations to complete.
The total training time was 11.80 hours ($42463$s).
From the result we can see that, \Name scales well when used for generating batches fixes for multiple vulnerabilities emerged in the same day.
Comparing with 1-Fix, batched 10-fixes only take 2.2X of training time instead of 10X, batched 22-fixes only take 3.5X of training time instead of 22X.
Also, \Name's training time for all 22 families of vulnerabilities (each with one fix) is less than 0.5 day, which is much lower than the critical time threshold of 1 day for \nday vulnerabilities.

\textbf{Accumulative Downtime (ADT)}:
We systematically increase the number of vulnerable devices (each with one vulnerability) from 1 to 10000 and measure the accumulative downtime across all devices.
Table~\ref{tab:evaluation_downtime} shows the ADT for \manualpatch, \softwarepatch, \networkfiltering and \Name.
For \manualpatch and \networkfiltering, their manual process can hardly scale beyond 100 devices given the 10 security admin (which is already high in manual cost) constrains in our IRB-study.
\softwarepatch can automated by scripts but its ADT is already 18.2 hours (65521s) when device number scale to 1000, and cannot scale to 10000 given resource constrains in our experiment.
In comparison, \Name can reduce the ADT by at least 10X comparing with other approaches, and \Name can scale to 10000 vulnerable devices with 1.5h ADT (5486s), easily supporting a branch office with thousands of employees (suppose each employee correspond to less than 10 devices).
Note that the ADT is the sum of downtime of all devices, not every device.

\begin{figure}[t]
\centering
\includegraphics[width=0.45\textwidth]{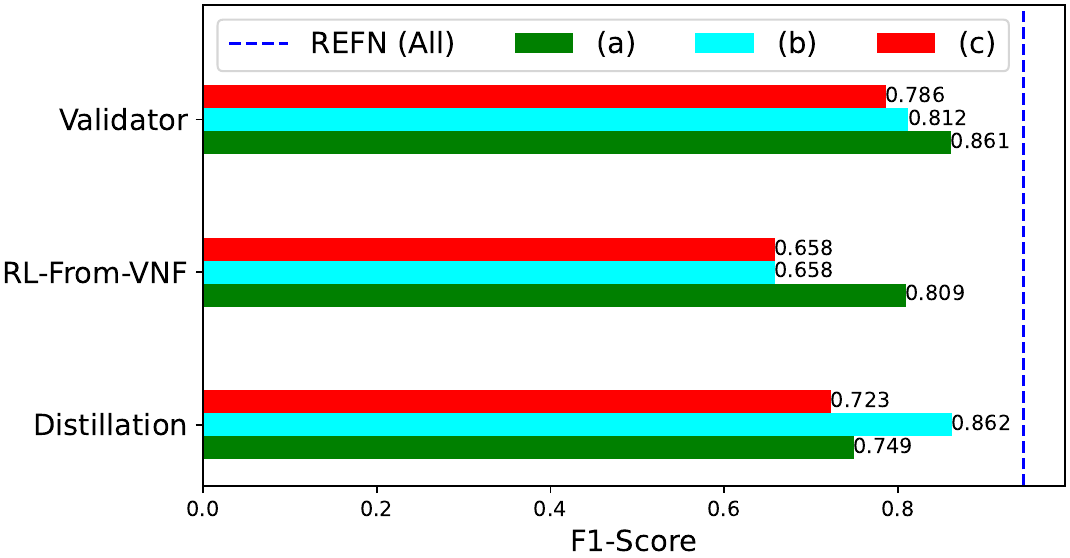}
\tightcaption{Ablation study on \Name components.}

\label{fig:evaluation_comp}
\end{figure}

\tightsubsection{Ablation Study}

As shown in Figure~\ref{fig:evaluation_comp}, we perform the ablation study and evaluate the benefit introduced by each of \Name's component with the following settings:
Distillation(a)-without agents;
Distillation(b)-without RAG;
Distillation(c)-remove all (manual text only);
RL-From-VNF(a)-without VNF-reward function;
RL-From-VNF(b)-without VNF-GRPO;
RL-From-VNF(c)-remove all (SFT only);
Validator(a)-without fuzzing \& trimming;
Validator(b)-without NToT;
Validator(c)-remove all (no validation, random reward).
Figure~\ref{fig:evaluation_comp} shows the F1-Score for each setting.
Without the \Distillation (Distillation(c)), the F1-Score (0.723) is lower than \Name (blue line) by 23.5\%.
This is because the training data and knowledge is the key for the RL process.
More specifically, the effectiveness impact of removing the Agent (Distillation(a), F1-Score 0.749, 20.7\% reduction) is greater than
removing the RAG (Distillation(b)).
When replacing the \RLVNF (RL-From-VNF(c)) with common SFT (Supervised Finetuning), the F1-Score (0.658) is lower than \Name by 30.4\%.
More specifically, the VNF-GRPO algorithm is absolutely necessary and without it (RL-From-VNF(b)), the F1-Score (0.658) is equally bad as removing the whole part.
Without the \Validator (Validator(c), random rewards), the F1-Score (0.786) is lower than \Name by 16.8\%.
Even though random rewards still demonstrates some effectiveness as it facilitates multiple iterations of training, 
the validation process as well as the fuzzing \& trimming (Validator(a)) and the NToT (Validator(b)) is still important as it further improves the F1-Score.
This is because the \Validator is key to ensure the fix is correct and reduce the error-susceptability.

\tightsection{Discussion}
\label{sec:discussion}


\noindent\textbf{Relation with traditional host-based patching}:
REFN is not designed to replace traditional host-based patching, but rather to complement it by providing rapid edge protection. 
The framework generates and deploys vulnerability-fixing filters rapidly to prevent large-scale exploitation during the critical window before host-based patches can be applied. This approach is particularly valuable for protecting legacy/embedded devices where patching is prohibitively difficult or costly.



\noindent\textbf{Handling encrypted traffic}:
Currently, \Name relies on the edge security gateway decryption (common in business/employee networks, e.g., Cisco Meraki scenarios) or context-inference mechanism to handle the exploitation via encrypted traffic.
In the future, we will explore enhanced methods including enterprise proxy integration and zero-trust authentication solutions to strengthen encrypted threat prevention.


\noindent\textbf{Handling LLM-based exploit tools}:
Theoretically, REFN can counter LLM-powered attack tools (e.g., HackerGPT~\cite{hackergpt}, WormGPT~\cite{wormgpt}) through knowledge distillation that extracts exploit patterns from these adversarial systems. By analyzing outputs from tools like WormGPT, REFN's distillation pipeline could preemptively identify and block novel attack vectors generated by malicious LLMs. Practical validation of this capability remains future work, with planned testing in real-world attack scenarios.


\tightsection{Conclusion}
\label{sec:conclusion}


The \nday vulnerabilities pose severe threats to diverse networked devices at massive scales. To combat this challenge, we introduce REFN, a novel framework that provide network-driven Reinforcement Learning to train LLMs and automatically generate and deploy vulnerability-fixing filters at the edge.
REFN effectively addresses large-scale exploitation across heterogeneous environments, demonstrating exceptional efficiency and scalability. Looking forward, \Name serves as an initial step toward rapidly preventing massive-scale exploitations at the edge.



%



\newpage
{
\section{Ethics considerations}
\label{sec:irb}

In this research, we conduct an IRB-approved study and invited ten security personal
with vulnerability mining competition experiences to perform three tasks.
The first task is a manual patch experiment, which requires the security admins to manually patch the vulnerable devices, using any available official documents and websites of the devices as the patching guide.
The second task is writing common network filter rules, which requires the security admins to manually generate the vulnerability fixing rules on top of basic prevention rules in Snort 2.9.8.0. 
The third task is to use common LLMs to generate and deploy patches and network filter rules, which requires the security admins to manually craft the LLM prompts and perform filter deployments.
Our Institutional Review Board (IRB) have censored the above evaluation and concluded that \textit{human subjects are not evolved} (because for any data used in this study, all the sensitive information including the personal's identity have been removed) and the highest ethical standards are met.
The experiment of this research is conducted in securely contained environment that satisfies the highest ethic standards.



}

\bibliographystyle{abbrv}

\bibliography{bibs/reft}


\newpage
{
\appendix

\section{Appendices}



\subsection{Evaluation Testbed}
\label{sec:appendix_evaluation_testbed}

\begin{figure}[h]
\centering
\includegraphics[width=0.45\textwidth]{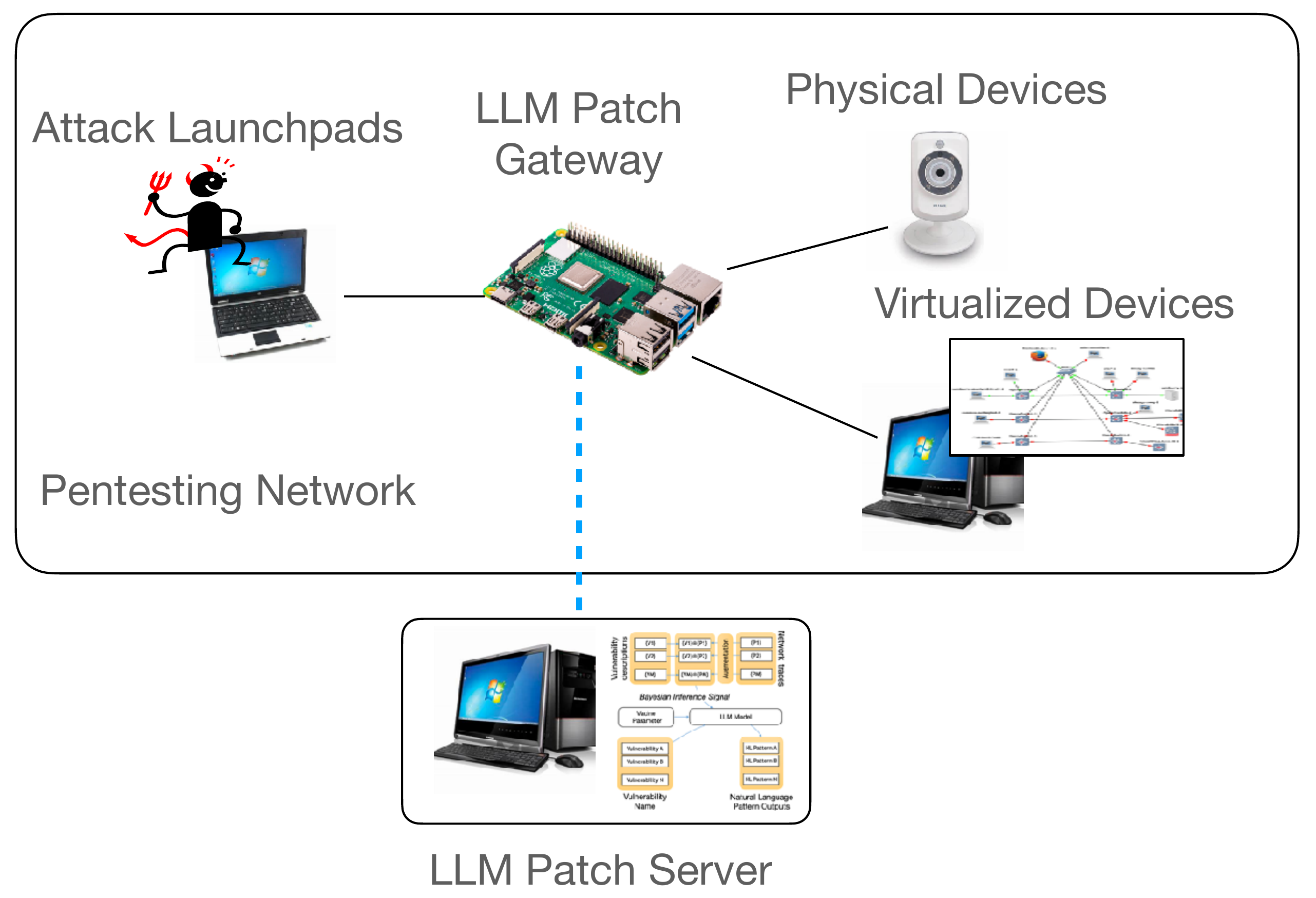}
\tightcaption{Evaluation Testbed.}
\label{fig:evaluation_topo}
\end{figure}

\subsection{OTNW ICS Network.}
\label{sec:appendix_evaluation_otnw}
\begin{figure}[h]
\vspace{-5pt}
\centering
\includegraphics[width=0.95\linewidth]{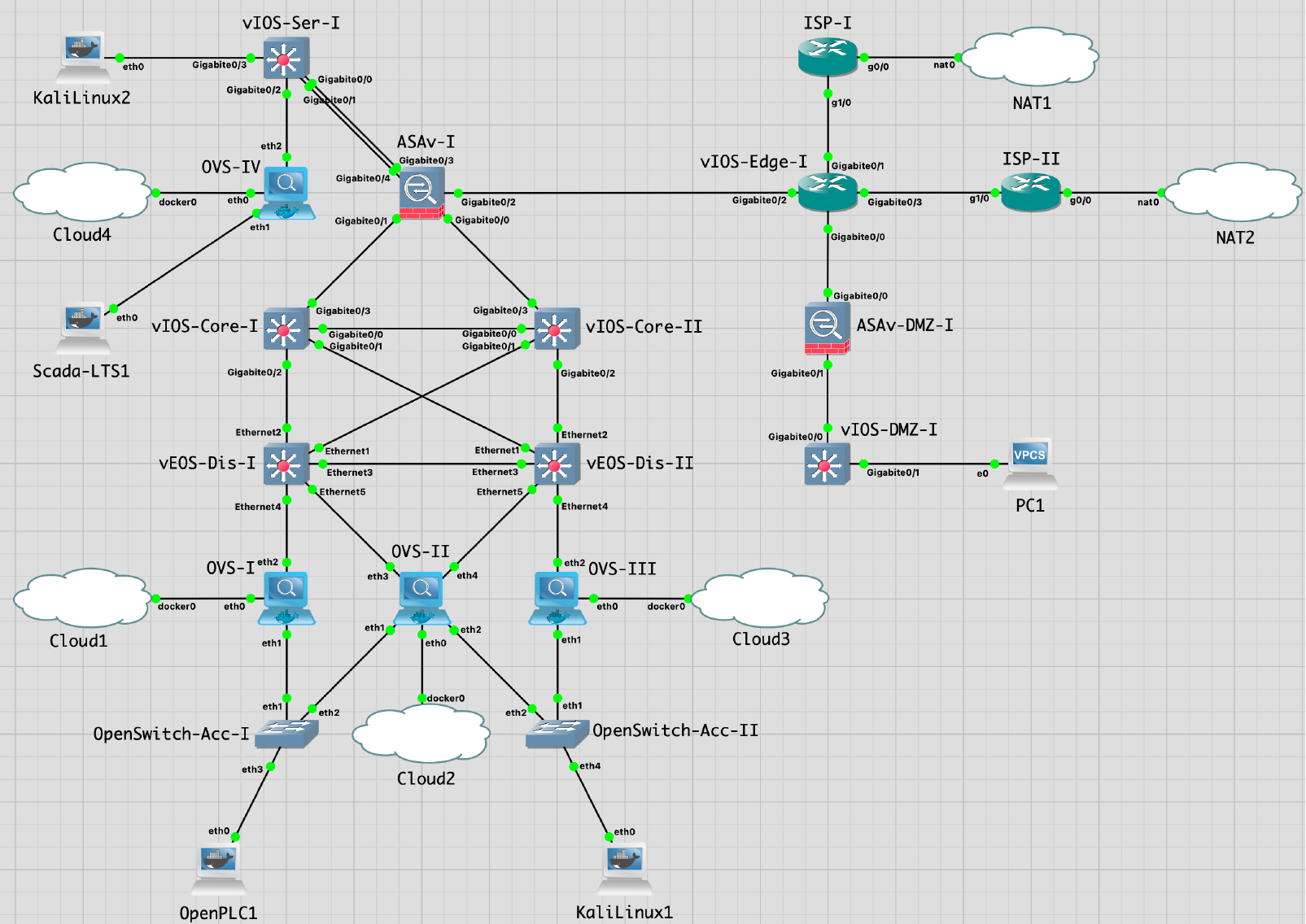}
\caption{OTNW ICS network topology.}
\vspace{-5pt}
\label{fig:evaluation_scalability_example_ics}
\end{figure}




\subsection{Patch Management Software List}
\label{sec:appendix_softwarepatch_list}

\begin{table}[h]
\begin{tabular}{ p{3.5cm} | p{3.5cm} }
\hline
Name & Version \\ 
\hline
ManageEngine Patch Manager Plus & 10.1.2220.20 \\ 
\hline
Chocolatey & 2.2.0 \\
\hline
Avira & 1.1.92.6 \\
\hline
Ninite & d0021\\
\hline
Patch My PC Home Updater & 4.5.0.3 \\
\hline
SUMo & 5.17.9.541 \\
\hline
Heimdal Free & 3.6.4 \\
\hline
Npackd & 1.26.9.0 \\
\hline
RuckZuck & 1.7.3.1 \\

\hline
\end{tabular}
\tightcaption{Patch management software.}
\label{tab:patch_management}
\end{table}

}
\end{document}